\documentclass[letterpaper]{article} 
\usepackage{aaai25}  
\usepackage{times}  
\usepackage{helvet}  
\usepackage{courier}  
\usepackage[hyphens]{url}  
\usepackage{graphicx} 
\usepackage{natbib}  
\usepackage{caption} 
\usepackage{algorithm}
\usepackage{algorithmicx}
\usepackage[noEnd=true]{algpseudocodex} 

\usepackage{newfloat}
\usepackage{listings}
\DeclareCaptionStyle{ruled}{labelfont=normalfont,labelsep=colon,strut=off} 
\floatstyle{ruled}
\newfloat{listing}{tb}{lst}{}
\floatname{listing}{Listing}
\usepackage{bm}
\usepackage{datatool}
\usepackage{fp}

\usepackage{amsmath}
\usepackage{amsthm}
\usepackage{amssymb}
\usepackage{mathtools}
\usepackage{stmaryrd}

\newtheorem{theorem}{Theorem}

\newtheorem{definition}{Definition}

\usepackage{ifthen}
\usepackage{enumitem}
\newlist{enuminline}{enumerate*}{2}
\setlist[enuminline,1]{label=(\arabic*),itemjoin={;\ }, itemjoin*={; and\ },after={.}}
\setlist[enuminline,2]{label=(\alph*),itemjoin={,\ }, itemjoin*={, and\ },after={}}
\newlist{iteminline}{itemize*}{2}
\setlist[iteminline]{label={},itemjoin={,\ }, itemjoin*={, and\ },afterlabel={},after={.}}
\setlist[iteminline,2]{after={},itemjoin*={,\ }}

\makeatletter
\newcommand{\thefontsize}{\f@size pt}
\makeatother
\newcommand*{\smallfont}{\fontsize{8}{9}\selectfont}
\newenvironment{sizeddisplay}
 {\par\nobreak\smallfont\noindent\ignorespaces}
 {\ignorespacesafterend}

\newcommand*{\textcite}[1]{\citeauthor{#1}~\shortcite{#1}}

\usepackage{xparse}

\usepackage[createShortEnv,conf={restate, no link to proof,text proof={Proof}}]{proof-at-the-end}

\newboolean{techreport}
\setboolean{techreport}{true}
\newboolean{reprochecklist}
\setboolean{reprochecklist}{false}

\usepackage{acronym}
\acrodef{BAT}{basic action theory}
\acrodef{CBAT}[C\textsuperscript{2}-BAT]{C\textsuperscript{2} basic action theory}
\acrodefplural{CBAT}[CBATs]{C\textsuperscript{2} basic action theories}
\acrodefplural{BAT}[BATs]{basic action theories}
\acrodef{SSA}{successor state axiom}
\acrodef{LTL}{Linear Temporal Logic}
\acrodefindefinite{LTL}{an}{a}
\acrodef{FOND}{fully observable nondeterministic}
\acrodef{LTLf}[LTL\textsubscript{f}]{LTL on finite traces}
\acrodefindefinite{LTLf}{an}{an}
\acrodef{NNF}{Negated Normal Form}
\acrodef{TNF}{Tail Normal Form}
\acrodef{XNF}{neXt Normal Form}
\acrodefindefinite{XNF}{an}{a}

\usepackage{listings}

\newcommand*{\ctwo}{C\textsuperscript{2}\xspace}
\newcommand*{\es}{\ensuremath{\mathcal{E \negthinspace S}}\xspace}
\newcommand*{\stdnames}[1][]{\ensuremath{\mathcal{N}\ifthenelse{\equal{#1}{}}{}{_{#1}}}\xspace}
\newcommand*{\primformulas}{\ensuremath{\mathcal{P}_F}\xspace}
\newcommand*{\traces}{\ensuremath{\mathcal{Z}}\xspace}
\newcommand*{\golog}{\textsc{Golog}\xspace}

\DeclareMathOperator{\lnext}{\mathcal{X}}
\DeclareMathOperator{\lwnext}{\mathcal{N}}
\DeclareMathOperator{\lfin}{\mathcal{F}}
\DeclareMathOperator{\lglob}{\mathcal{G}}
\newcommand*{\luntil}{\ensuremath{\mathbin{\mathcal{U}}}\xspace}
\newcommand*{\lduntil}{\ensuremath{\mathbin{\mathcal{R}}}\xspace}
\newcommand*{\lrelease}{\lduntil}
\newcommand*{\ltail}{\ensuremath{\mi{Tail}}\xspace}
\newcommand*{\worlds}{\ensuremath{\mathcal{W}}\xspace}
\newcommand*{\bat}{\ensuremath{\mathcal{D}}\xspace}
\newcommand*{\prog}{\ensuremath{\mathcal{G}}\xspace}
\newcommand*{\depgraph}[1][\bat]{\ensuremath{\Delta_{#1}}\xspace}
\DeclareMathOperator{\fd}{fd}
\DeclareMathOperator*{\final}{Fin}
\newcommand*{\chargraph}[1][\delta]{\ensuremath{\mathcal{C}_{#1}}\xspace}
\newcommand*{\actions}{\ensuremath{\mathcal{A}}\xspace}
\newcommand*{\context}{\ensuremath{\mathcal{C}}\xspace}
\newcommand*{\effects}[1][\bat,\actions]{\ensuremath{\mathfrak{E}^{#1}}\xspace}
\newcommand*{\effdesc}{\ensuremath{\varepsilon}\xspace}
\newcommand*{\contcon}{\ensuremath{\kappa}\xspace}
\NewDocumentCommand{\ts}{O{\mathcal{G}} O{\Phi}}{\ensuremath{\mathbb{A}_{#1}^{#2}}\xspace}
\newcommand*{\tsstates}{\ensuremath{\mathcal{S}}\xspace}

\DeclareMathOperator{\sub}{sub}
\DeclareMathOperator{\tnf}{tnf}
\DeclareMathOperator{\tnft}{t}
\DeclareMathOperator{\xnf}{xnf}
\DeclareMathOperator{\cl}{cl}
\DeclareMathOperator{\types}{Types}
\DeclareMathOperator{\type}{type}
\DeclareMathOperator{\typeelements}{TE}
\DeclareMathOperator{\plays}{plays}
\DeclareMathOperator{\closure}{cl}
\DeclareMathOperator{\propatoms}{PA}
\DeclareMathOperator{\nilop}{nil}
\DeclareMathOperator{\playacts}{Acts}
\newcommand*{\nil}{\ensuremath{\nilop}\xspace}

\DeclareMathOperator*{\poss}{Poss}

\newcommand*{\rfluent}[1]{\ensuremath{\xcapitalisewords{\mathit{#1}}}\xspace}
\newcommand*{\action}[2]{\ensuremath{\mathit{#1}}(#2)\xspace}
\newcommand*{\paction}[2]{\ensuremath{\underline{\action{#1}{#2}}}\xspace}

\newcommand*{\eqdef}{\mathrel{\dot{=}}}
\newcommand*{\la}{\langle}
\newcommand*{\ra}{\rangle}
\newcommand*{\reg}{\ensuremath{\mathcal{R}}\xspace}
\newcommand*{\mi}[1]{\ensuremath{\mathit{#1}}\xspace}

\gdef\M#1{\ensuremath{#1}}

\newcommand{\seq}{\M{;}}
\newcommand{\nondet}{\M{|}}

\newcommand{\iter}[1]{\M{{#1}^*}}

\newcommand{\conc}{\M{|\!|}}

\newcommand{\effset}[1]{\ensuremath{{\mathfrak{E}^{\D\negthinspace,\Act}_{#1}}}\xspace}
\newcommand{\adddeleff}[2]{\ensuremath{\langle #1^{\pm},#2 \rangle}\xspace}
\newcommand{\fdfluent}[1]{\ensuremath{\mathsf{fd}_{\D}(#1)}\xspace}
\newcommand{\effneg}[1]{\ensuremath{\mathsf{eff}^{-}_{\Act}(#1)}\xspace}
\newcommand{\effpos}[1]{\ensuremath{\mathsf{eff}^{+}_{\Act}(#1)}\xspace}
\newcommand{\addeff}[2]{\ensuremath{\langle #1^{+},#2 \rangle}\xspace}
\newcommand{\deleff}[2]{\ensuremath{\langle #1^{-},#2 \rangle}\xspace}
\newcommand{\Regress}[2]{\ensuremath{\mathcal{R}[#1,#2]}\xspace}
\newcommand{\E}{\ensuremath{\mathsf{E}}\xspace}
\newcommand{\D}{\ensuremath{\mathcal{D}}\xspace}

\newcommand{\Act}{\ensuremath{\mathcal{A}}\xspace}
\newcommand{\fdbat}{\ensuremath{\mathsf{fd}(\D)}\xspace}
\newcommand{\effform}{\effdesc}
\newcommand{\effcon}{\contcon}
\newcommand{\bigldo}{\mathchoice{\bigl(}{(}{(}{(}}
\newcommand{\bigldc}{\mathchoice{\bigr)}{)}{)}{)}}
\newcommand{\gammap}{\ensuremath{\bigldo{\gamma^+_F}\bigldc^a_\alpha}\xspace}
\newcommand{\gamman}{\ensuremath{\bigldo{\gamma^-_F}\bigldc^a_\alpha}\xspace}
\newcommand{\effsetcomplete}{\ensuremath{{\mathfrak{E}^{\D\negthinspace,\Act}}}\xspace}

\newcommand{\eseq}{\la\ra}
\newcommand{\conf}[2]{\la #1,#2 \ra}

\algblockdefx[Name]{Pick}{EndPick}[2]{$\mathbf{\pi}$ #1 \::\: #2}{}

\newcommand{\termcond}[1]{\M{\varphi(#1)}}
\newcommand{\progtrans}[2]{\M{\xrightarrow{#1:#2}}}

\newcommand{\true}{\M{\top}}
\newcommand{\false}{\M{\bot}}

\newcommand*{\vergo}{\texttt{vergo}\xspace}

\DeclareMathOperator*{\Succ}{Succ}

\usepackage[textsize=scriptsize,disable]{todonotes}
\usepackage{mfirstuc}
\usepackage{xspace}

\title{\acs*{LTLf} Synthesis on First-Order Agent Programs in Nondeterministic Environments}
\author{
  Till Hofmann\textsuperscript{\rm 1},
  Jens Claßen\textsuperscript{\rm 2}
}
\affiliations{
\textsuperscript{\rm 1}Department of Computer Science, RWTH Aachen University\\
\textsuperscript{\rm 2}Institute for People and Technology, Roskilde University\\
till.hofmann@cs.rwth-aachen.de,
classen@ruc.dk
}

\begin{document}

\maketitle

\begin{abstract}
  We investigate the synthesis of policies for high-level agent programs expressed in Golog, a language based on situation calculus that incorporates nondeterministic programming constructs. Unlike traditional approaches for program realization that assume full agent control or rely on incremental search, we address scenarios where environmental nondeterminism significantly influences program outcomes. Our synthesis problem involves deriving a policy that successfully realizes a given Golog program while ensuring the satisfaction of a temporal specification, expressed in Linear Temporal Logic on finite traces (LTLf), across all possible environmental behaviors. By leveraging an expressive class of first-order action theories, we construct a finite game arena that encapsulates program executions and tracks the satisfaction of the temporal goal. A game-theoretic approach is employed to derive such a policy. Experimental results demonstrate this approach's feasibility in domains with unbounded objects and non-local effects. This work bridges agent programming and temporal logic synthesis, providing a framework for robust agent behavior in nondeterministic environments.

\end{abstract}

\section{Introduction}

Agents operating in dynamic environments often need to react to changes beyond their control.
For example, a service robot may be tasked with serving coffee to customers, who may place an order at any time.
Also, some actions may have unexpected outcomes, e.g., while attempting to fulfill the task, the robot might accidentally drop the coffee.
Such scenarios can be modeled as \acfi{FOND} planning tasks~\cite{geffnerConciseIntroductionModels2013,ghallabAutomatedPlanningActing2016} where actions have multiple possible outcomes, or as reactive synthesis problem (e.g., based on \acfi{LTLf}~\cite{degiacomoSynthesisLTLLDL2015}), where certain propositions are controlled by the agent while others are governed by the environment.
However, these approaches have some limitations.
They assume a fixed, finite set of propositions, effectively imposing a closed-world assumption and requiring a completely known initial state.
Furthermore, they rely on alternating actions between the agent and the environment, which fails to capture scenarios where the environment may perform an arbitrary number of actions before the agent can respond.
Additionally, existing solutions often generate arbitrary plans or policies without incorporating user-specified partial strategies unless explicitly encoded in the specification.

On the other hand, \golog~\cite{levesqueGOLOGLogicProgramming1997}, a well-established agent programming language, offers significant flexibility.
Based on the situation calculus~\cite{mccarthyPhilosophicalProblemsStandpoint1969,reiterKnowledgeActionLogical2001}, \golog supports first-order reasoning over arbitrarily large or even infinite domains and accommodates incomplete information about the initial state.
It also allows for the specification of partial strategies through nondeterministic programs.
Given such a program, \emph{program realization} is the task of resolving program nondeterminism to produce a successful program execution, e.g., by means of search, or in an online incremental fashion~\cite{degiacomoIndiGologHighlevelProgramming2009}.
However, it is typically assumed that the agent is in complete control, even if it only has incomplete knowledge~\cite{reiterKnowledgebasedProgrammingSensing2001,classenKnowledgebasedProgramsDefaults2016} or its actions are stochastic~\cite{boutilierDecisiontheoreticHighlevelAgent2000}.
Recently, the situation calculus has been extended with nondeterministic actions~\cite{degiacomoNondeterministicSituationCalculus2021,classenAccountIntensionalExtensional2021} similar to FOND planning, where the environment chooses an outcome.
However, this still assumes that agent and environment act in turns.

To address these limitations, we propose an extension to \golog that partitions actions into agent actions and environment actions.
In this framework, the agent selects among currently applicable agent actions, guided by the program and the basic action theory, but cannot constrain the environment.
The environment may select any applicable environment action or any action chosen by the agent.
This allows for arbitrary sequences of environment actions, similar to the \emph{supervisory control} paradigm~\cite{ramadgeControlDiscreteEvent1989}.
We also propose to describe the agent's goal as a temporal formula, which allows for formulating trajectory constraints such as safety and reachability on the program.

In this setting, program realization becomes a synthesis task.
Given a \golog program and a temporal goal, the task is to synthesize a policy that executes the program while satisfying the temporal goal, independent of and reacting to all possible environment behaviors.
In this paper, we focus on the decidable fragment of \golog with acyclic \aclp{BAT} restricted to \ctwo \cite{zarriessDecidableVerificationGolog2016} and temporal goals given as \acs{LTLf}\acused{LTLf} formulas~\cite{degiacomoLinearTemporalLogic2013}.
We provide a decidable approach for this problem by constructing a finite game arena that captures all possible program executions while tracking the satisfaction of the temporal specification, and then applying a game-theoretic approach to synthesize a policy.
Exploiting an encoding of \ac{LTLf} formulas that interprets temporal formulas as propositional atoms~\cite{liSATbasedExplicitLTLf2020}, the construction works on-the-fly and avoids building irrelevant parts.

The remainder of this paper is structured as follows.
After discussing related work in Section~\ref{sec:related}, we summarize \golog and introduce \ac{LTLf} in the context of \golog programs in Section~\ref{sec:preliminaries}.
We describe the synthesis approach in Section~\ref{sec:approach} and evaluate it experimentally in Section~\ref{sec:evaluation}, before concluding in Section~\ref{sec:conclusion}.

\section{Related Work}
\label{sec:related}


Verification of \golog programs has been explored in various contexts.
Initially, verification efforts relied on manual proofs~\cite{degiacomoNonterminatingProcessesSituation1997,liuHoarestyleProofSystem2002,shapiroCognitiveAgentsSpecification2002}.
\textcite{classenLogicNonterminatingGolog2008} describe a (possibly not terminating) system that is capable of automatically verifying properties of non-terminating \golog programs.
Later research identified decidable fragments of \golog grounded in \ctwo, the decidable two-variable fragment of first-order logic with counting~\cite{gradelTwovariableLogicCounting1997}.
Verification of \golog programs with \emph{context-free} or \emph{local-effect} \acp{BAT} in \ctwo and with pick operators restricted to finite domains is decidable for properties in CTL~\cite{classenExploringBoundariesDecidable2014}, LTL~\cite{zarriessDecidabilityVerifyingLTL2014}, and CTL\textsuperscript{*}~\cite{zarriessVerifyingCTLProperties2014}.
Beyond local-effect \acp{BAT}, verification remains decidable if the \ac{BAT} is \emph{acyclic}, i.e., there is no cyclic dependency between fluents in the effect descriptors, or \emph{flat}, i.e., effect descriptors are quantifier-free~\cite{zarriessDecidableVerificationGolog2016}.
\emph{Bounded theories}, where the number of objects described by any situation is bounded, also results in decidable verification
\cite{degiacomoBoundedSituationCalculus2016}.
All these approaches rely on a finite abstraction of the infinite program configuration space, which yields decidability, and hence could be used as basis for our approach.

Related to verification is \emph{synthesis} of temporal properties, which can be described as two-player games between the system and the environment~\cite{abadiRealizableUnrealizableSpecifications1989,pnueliSynthesisReactiveModule1989}.
Given a specification, e.g., in \ac{LTL}, and a partition of the symbols into controllable and uncontrollable ones, the players alternate selecting a subset of their symbols.
%
\ac{LTL} has also been used to describe temporally extended goals for planning~\cite{bacchusPlanningTemporallyExtended1998,degiacomoAutomataTheoreticApproachPlanning2000,geffnerConciseIntroductionModels2013}, possibly resulting in infinite plans~\cite{patriziComputingInfinitePlans2011}.
\ac{LTL} can also be used to specify \emph{conformant planning} problems with temporally extended goals~\cite{calvaneseReasoningActionsPlanning2002}
and synthesis is related to FOND planning~\cite{camachoNonDeterministicPlanningTemporally2017,camachoFiniteLTLSynthesis2018,degiacomoAutomatatheoreticFoundationsFOND2018} as a nondeterministic effect can be seen as an environment action.
Moreover, there has been a particular interest in \ac{LTLf}~\cite{degiacomoLinearTemporalLogic2013}, where the synthesis problem can be solved by transforming the \ac{LTLf} specification into a finite automaton~\cite{degiacomoSynthesisLTLLDL2015}.
Like \ac{LTL}, \ac{LTLf} synthesis is \textsc{2ExpTime}-complete, although \ac{LTLf} synthesis tools usually perform better.
Recently, several methods have been proposed to improve the performance of \ac{LTLf} synthesis, e.g., based on BDDs~\cite{zhuSymbolicLTLfSynthesis2017} and on-the-fly forward search~\cite{xiaoOntheflySynthesisLTL2021,degiacomoLTLfSynthesisGraph2022,favoritoEfficientAlgorithmsLTLf2023}.

\section{Preliminaries}
\label{sec:preliminaries}

We describe the logic \es and an \es-based variant of \golog and then introduce \ac{LTLf} in the context of \golog programs.

\subsection{The Logic \es}

The logic \es\ \cite{lakemeyer2010semantic} is a first-order modal variant of the situation calculus.
Following \cite{zarriessDecidableVerificationGolog2016}, we consider \es\ formulas restricted to \ctwo.

\paragraph{Syntax}

{\em Terms} are of sort \emph{object} or \emph{action}.
We use $x,y,\ldots$ (possibly with decorations) to denote object variables, and $a$ for a variable of sort action.
$N_O$ is a countably infinite set of {\em object constant symbols},
and $N_A$ a countably infinite set of action function symbols whose arguments are all of sort object.
Let $\stdnames[O]$ denote the set of all ground terms (called {\em standard names}) of sort object,
and $\stdnames[A]$ those of sort action.
Formulas are constructed over equality atoms and {\em fluent} predicates
with at most two arguments of sort object,
using the usual Boolean connectives, quantifiers, counting quantifiers,\todo{Do we actually support them?}
as well as modalities $\Box \phi$ (``$\phi$ holds after any sequence of actions''),
and $[t]\phi$ (``$\phi$ holds after executing action $t$'').
We call a formula {\em fluent} if it does not mention $\Box$ or $[\cdot]$.
A \emph{sentence} is a formula without free variables.  
A \emph{\ctwo-fluent formula} is a fluent formula without actions and with at most two variables.

\paragraph{Semantics}

A {\em trace} is a finite sequence of action standard names.
When a trace represents a history of already executed actions, it is called a {\em situation}.
For a trace $z = \la \alpha_1, \ldots, \alpha_n \ra \in \traces$, we write $|z|$ for the length $n$ of $z$, $z \cdot \alpha$ for the concatenation $\la \alpha_1, \ldots, \alpha_n, \alpha \ra$ of $z$ with an action $\alpha$, $z[i]$ for the $i$th action $\alpha_i$, $z[..i]$ for the prefix $\la \alpha_1, \ldots, \alpha_i \ra$, and $z[i..]$ for the suffix $\la \alpha_i, \ldots, \alpha_n \ra$.
Let $\traces = \stdnames[A]^*$ be the set of all traces, and $\primformulas$ the set of all {\em primitive formulas} $F(n_1,...,n_k)$,
where $F$ is a $k$-ary fluent with $0 \leq k \leq 2$ and the $n_i$ are object standard names.
A \emph{world} $w$ maps primitive formulas and situations to truth values, i.e.,
$ w : \primformulas \times \traces \rightarrow \{0,1\}$.
The set of all worlds is denoted by $\worlds$.
\begin{definition}[Truth of Formulas]
  Let $w \in \worlds$ be a world and $\alpha$ an action standard name.
  We define for every $z \in \traces$:
  \begin{enumerate}
    \item $w, z \models F(n_1, \ldots, n_k)$ iff $w[F(n_1, \ldots, n_k),z] = 1$;
    \item $w, z \models (n_1 = n_2)$ iff $n_1$ and $n_2$ are identical;
    \item $w, z \models \phi_1 \wedge \phi_2$ iff $w, z \models \phi_1$ and $w, z \models \phi_2$;
    \item $w, z \models \neg \phi$ iff $w, z \not\models \phi$;
    \item $w, z \models \forall x. \phi$ iff $w, z \models \phi^x_n$ for every $n \in \stdnames[x]$;
    \item $w, z \models \exists^{\leq m} x.\phi$ iff $\lvert \{ n \in \stdnames[x] \mid w, z \models \phi^x_n \} \rvert \leq m$;
    \item $w, z \models \exists^{\geq m} x.\phi$ iff $\lvert \{ n \in \stdnames[x] \mid w, z \models \phi^x_n \} \rvert \geq m$;
    \item $w, z \models \square \phi$ iff $w, z \cdot z' \models \phi$ for every $z \in \traces$;
    \item $w, z \models [\alpha] \phi$ iff $w, z \cdot \alpha \models \phi$.
  \end{enumerate}
\end{definition}
  Here, $\stdnames[x]$ refers to the set of all standard names of the same sort as $x$,
  and $\phi^x_n$ the result of simultaneously replacing all free occurrences of $x$ in $\phi$ by $n$.
  We understand $\lor$, $\exists$, $\supset$, $\equiv$, $\top$ and $\bot$ as the usual abbreviations.
  For a set of sentences $\Sigma$ and a sentence $\alpha$,
  we write $\Sigma\models\alpha$ (read: $\Sigma$ entails $\alpha$) to mean that
  for every $w$, if $w,\eseq \models \alpha'$ for every $\alpha'\in\Sigma$, then $w,\eseq \models \alpha$.
  Finally, we write $\models \alpha$ (read: $\alpha$ is valid) to mean $\{\} \models \alpha$.
Note that rule 2 above includes a unique names assumption for actions and objects into the semantics.

\subsection{Basic Action Theories}

To encode a dynamic domain, we employ a \acfi{BAT} \cite{reiterKnowledgeActionLogical2001} with additional restrictions \cite{zarriessDecidableVerificationGolog2016} for ensuring decidability:
\begin{definition}[Basic Action Theory]
  \label{def:BAT}
  A \acfi{BAT} $\bat = \bat_0 \cup \bat_\text{post}$ is a set of axioms, where
  $\bat_0$ is a finite set of \ctwo-fluent sentences describing the initial state of the world,
      and $\bat_\text{post}$ is a finite set of \acfp{SSA}, one for each fluent, of the form\footnote{
        The operator $\square$ has lowest precedence while $[\cdot]$ has highest precedence and free variables are implicitly assumed to be universally quantified from the outside.
      }
      $\square [a] F(\vec{x}) \equiv \gamma_F^+ \vee F(\vec{x}) \wedge \neg \gamma_F^-$,
      where the \emph{positive effect condition $\gamma_F^+$} and the \emph{negative effect condition $\gamma_F^-$} are disjunctions of formulas of the form $\exists \vec{y}. \left(a = A(\vec{v}) \wedge \effdesc \wedge \contcon\right)$ such that
      \begin{itemize}
        \item the free variables of the formula $\exists \vec{y}. \left(a = A(\vec{v}) \wedge \effdesc \wedge \contcon\right)$ are among $\vec{x}$ and $a$,
        \item $A(\vec{v})$ is an action term and $\vec{v}$ contains $\vec{y}$,
        \item the \emph{effect descriptor} \effdesc is a fluent formula with no terms of sort action and the number of variables in \effdesc that do not occur in $\vec{v}$ or occur bound in \effdesc is less than or equal to two,
        \item the \emph{context condition} \contcon is a fluent formula with free variables among $\vec{v}$, no terms of sort action, and at most two bound variables.
      \end{itemize}
\end{definition}
\noindent
Intuitively, the effect descriptor is the part of the effect condition that expresses {\em which objects} are affected,
while the context condition encodes {\em whether} the effect takes place.

\subsubsection{Acyclic \acp{BAT}}
For a \ac{BAT} \bat, we can construct the \emph{fluent dependency graph} \depgraph, which captures the dependencies between fluents in the effect descriptors.
In \depgraph, each node is a fluent of \bat and there is a directed edge $(F, F')$ from fluent $F$ to fluent $F'$ if there exists a disjunct $\exists \vec{y}.(a = A(\vec{v}) \wedge \effdesc \wedge \contcon)$ in $\gamma_F^+$ or $\gamma_F^-$ such that $F'$ occurs in \effdesc.
A \ac{BAT} is \emph{acyclic} if \depgraph is acyclic.
Furthermore, the \emph{fluent depth} of an acyclic \ac{BAT}, denoted by $\fd(\bat)$, is the length of the longest path in \depgraph and the \emph{fluent depth of  $F$ w.r.t.\ \bat}, denoted by $\fd_\bat(F)$, is the length of the longest path in \depgraph starting in $F$.

\subsection{\golog Programs}


We consider a set of program expressions that includes
ground actions ($\alpha$),
tests for \ctwo-fluent sentences ($\phi?$),
sequence of subprograms ($\delta_1\seq\delta_2$),
nondeterministic choice ($\delta_1\nondet\delta_2$),
interleaved concurrent execution ($\delta_1\conc\delta_2$),
and nondeterministic iteration ($\iter{\delta}$).
We write $\nil \eqdef \top?$ for the empty program that always succeeds.

A \emph{\golog program} $\prog = (\bat, \delta)$ consists of a \ctwo-\ac{BAT} $\bat = \bat_0 \cup \bat_\text{post}$ and a program expression $\delta$, where all fluents occurring in \bat and $\delta$ have \iac{SSA} in $\bat_\text{post}$.
For a program $\prog = (\bat, \delta)$, we write $\actions_\prog$ for all action terms occurring in $\delta$ and we may omit the subscript if \prog is clear from context.

The semantics of \golog programs is based on transitions between configurations, where a configuration $\la z, \rho \ra$ consists of a sequence of already performed actions $z \in \traces$ and the remaining program $\rho \in \sub(\delta)$.
Given a world $w \in \worlds$, the transition relation $\xrightarrow{w}$ among configurations is defined inductively.
The set of final configurations $\final(w)$ defines the configurations where the program may terminate.  

\begin{definition}[Program Transition Semantics]
For any world $w$, the set of final configurations $\final(w)$ is the smallest set such that
\begin{sizeddisplay}
  \begin{align*}
    \conf{z}{\phi?}\in\final(w) &\text{ if } w,z \models \phi
    \\
    \conf{z}{\delta_1\seq\delta_2}\in\final(w) &\text{ if } \conf{z}{\delta_1}\in\final(w) \text{ and } \conf{z}{\delta_2}\in\final(w)
    \\
    \conf{z}{\delta_1\nondet\delta_2}\in\final(w) &\text{ if }\conf{z}{\delta_1}\in\final(w) \text{ or } \conf{z}{\delta_2}\in\final(w)
    \\
    \conf{z}{\delta_1\conc\delta_2}\in\final(w) &\text{ if } \conf{z}{\delta_1}\in\final(w) \text{ and } \conf{z}{\delta_2}\in\final(w)
    \\
    \conf{z}{\iter{\delta}}\in\final(w)
  \end{align*}
\end{sizeddisplay}
\label{def:ProgramTransitionSemantics}
For any world $w$, the transition relation $\xrightarrow{w}$ among configurations is the least set satisfying
\begin{sizeddisplay}
  \begin{align*}
    \conf{z}{\alpha} &\xrightarrow{w}  \conf{z \cdot \alpha}{\nil} \text{ if $\alpha$ is a ground action }
    \\
    \conf{z}{\delta_1\seq\delta_2} &\xrightarrow{w} \conf{z'}{\rho\seq\delta_2} \text{ if } \conf{z}{\delta_1} \xrightarrow{w} \conf{z'}{\rho}
    \\
    \conf{z}{\delta_1\seq\delta_2} &\xrightarrow{w} \conf{z'}{\rho} \text{ if }\conf{z}{\delta_1}\in\final(w) \text{ and } \conf{z}{\delta_2} \xrightarrow{w} \conf{z'}{\rho}
    \\
    \conf{z}{\delta_1\nondet\delta_2} &\xrightarrow{w} \conf{z'}{\rho} \text{ if } \conf{z}{\delta_1} \xrightarrow{w} \conf{z'}{\rho} \text{ or } \conf{z}{\delta_2} \xrightarrow{w} \conf{z'}{\rho}
    \\
    \conf{z}{\delta_1\conc\delta_2} &\xrightarrow{w} \conf{z'}{\rho\conc\delta_2} \text{ if } \conf{z}{\delta_1} \xrightarrow{w} \conf{z'}{\rho}
    \\
    \conf{z}{\delta_1\conc\delta_2} &\xrightarrow{w} \conf{z'}{\delta_1\conc\rho} \text{ if } \conf{z}{\delta_2} \xrightarrow{w} \conf{z'}{\rho}
    \\
    \conf{z}{\iter{\delta}} &\xrightarrow{w} \conf{z'}{\rho\seq\iter{\delta}} \text{ if } \conf{z}{\delta} \xrightarrow{w} \conf{z'}{\rho}
  \end{align*}
\end{sizeddisplay}
\end{definition}




We write $\|\delta\|^z_w$ for the set of traces starting in configuration $\la z, \delta \ra$ and ending in a final configuration.

\subsubsection{Situation-Determined Programs}
Following \cite{DBLP:conf/aamas/GiacomoLM12}, we say that
a program $\prog = (\bat, \delta)$ is {\em situation-determined}, iff
for all $w \in \worlds$ with $w \models \bat$, all $z,z' \in \traces$,
and all program expressions $\delta',\delta''$:
$\la z,\delta \ra \xrightarrow{w}^* \la z',\delta' \ra$ and
   $\la z,\delta \ra \xrightarrow{w}^* \la z',\delta'' \ra$ implies
   $\delta' = \delta''$.
We assume that all programs are situation-determined.

\subsection{\ac{LTLf}}
For temporal properties, we define temporal formulas with the same syntax as \ac{LTLf} formulas, but replacing propositions with \ctwo-fluent sentences $\phi$, i.e., $\Phi ::= \phi \mid \Phi \wedge \Phi \mid \lnext \Phi \mid \Phi \luntil \Phi$.
For a temporal formula $\Phi$, we denote the set of subformulas of $\Phi$ with $\closure(\Phi)$.
For a set of formulas $\Psi$, we write $\bigwedge \Psi$ for $\bigwedge_{\Phi \in \Psi} \Phi$.
As usual, we define $\lfin \Phi \eqdef \top \luntil \Phi$ and $\lglob \Phi \eqdef \neg \lfin \neg \Phi$,
as well as
$\Phi_1 \vee \Phi_2 \eqdef \neg (\neg \Phi_1 \wedge \neg \Phi_2)$, $\lwnext \Phi \eqdef \neg \lnext \neg \Phi$, and $\Phi_1 \lrelease \Phi_2 \eqdef \neg (\neg \Phi_1 \luntil \neg \Phi_2)$.
We define the truth of a temporal formula $\Phi$, given a world $w$ and traces $z, z'$:
\begin{itemize}
  \item $w, z, z' \models \phi$ iff $w, z \models \phi$,
  \item $w, z, z' \models \Phi_1 \wedge \Phi_2$ iff $w, z, z' \models \Phi_1$ and $w, z, z' \models \Phi_2$,
  \item $w, z, z' \models \lnext \Phi$ iff $z' = \alpha \cdot z'' \neq \la\ra$ and $w, z \cdot \alpha, z'' \models \Phi$,
  \item $w, z, z' \models \Phi_1 \luntil \Phi_2$ iff there exists $k \leq |z'|$ such that $w, z \cdot z'[..k], z'[k + 1..] \models \Phi_2$ and for all $0 \leq i < k$, $w, z \cdot z'[..i], z'[i + 1..] \models \Phi_1$.
\end{itemize}

\subsubsection{\acs*{TNF} and \acs*{XNF}}
As we intend to track the satisfiability of the temporal formula $\Phi$ over the traces of the program, we adapt \acf{TNF} and \acf{XNF} from \cite{liSATbasedExplicitLTLf2020}.
\ac{TNF} explicitly marks the end of satisfying traces, while
\ac{XNF} allows us to split the temporal formula into a local part, which can be evaluated at the current state, and a future part, which is evaluated against the remaining trace.
First, we say a formula is in \acfi{NNF} if all negations are in front of only atoms.
Each \ac{LTLf} formula can be transformed into \ac{NNF} by using the dual operators to push negation inwards.
Based on \ac{NNF}, we define \ac{TNF}, which marks the last state of satisfying traces:
\begin{definition}
  Let $\Phi$ be an \ac{LTLf} formula in \ac{NNF}.
  Its \ac{TNF} $\tnf(\Phi)$ is defined as $\tnft(\Phi) \wedge \lfin \ltail$, where \ltail is a new atom to identify the last state of satisfying traces and $\tnft(\Phi)$ is an \ac{LTLf} formula defined recursively as follows:
  \begin{enumerate}
    \item $\tnft(\Phi) = \Phi$ if $\Phi$ is $\top, \bot$, or a \ctwo-fluent sentence;
    \item $\tnft(\lnext(\Psi)) = \neg \ltail \wedge \lnext(\tnft(\Psi))$;
    \item $\tnft(\lwnext(\Psi)) = \ltail \vee \lnext(\tnft(\Psi))$;
    \item $\tnft(\Phi_1 \wedge \Phi_2) = \tnft(\Phi_1) \wedge \tnft(\Phi_2)$;
    \item $\tnft(\Phi_1 \vee \Phi_2) = \tnft(\Phi_1) \vee \tnft(\Phi_2)$;
    \item $\tnft(\Phi_1 \luntil  \Phi_2) = (\neg \ltail \wedge \tnft(\Phi_1)) \luntil \tnft(\Phi_2)$;
    \item $\tnft(\Phi_1 \lduntil \Phi_2) = (\ltail \vee \tnft(\Phi_1)) \lduntil \tnft(\Phi_2)$.
  \end{enumerate}
\end{definition}

When interpreting a \ac{TNF} formula over a trace, \ltail needs to be treated separately, as it is not a fluent sentence.
We define: 
$w, z, z' \models \ltail \text{ iff } z' = \la\ra$.
It can be shown that $\Phi$ and $\tnf(\Phi)$ are equivalent:\footnote{\ifthenelse{\boolean{techreport}}{Proofs can be found in the appendix.}{Proofs can be found in~\cite{hofmannLTLfSynthesisFirstOrder2024}.}}
\begin{theoremE}
  Let $\Phi$ be a temporal formula, $w$ a world, and $z$ and $z'$ traces.
  Then $w, z, z' \models \Phi$ iff $w, z, z' \models \tnf(\Phi)$.
\end{theoremE}
\begin{proofE}
  We show by structural induction on $\Phi$ that for arbitrary $w, z, z'$, it holds that $w, z, z' \models \Psi$ iff $w, z , z' \models \tnf(\Psi)$.
  \begin{itemize}
    \item Let $\Phi$ be $\top, \bot$, or a \ctwo-fluent sentence.
      Then  $\tnf(\Phi) = \Phi$ and the claim holds.
    \item The Boolean cases follow immediately by induction.
    \item Let $\Phi$ be $\lnext(\Psi)$.
      Then $w, z, z' \models \lnext(\Psi)$ iff $z' = \alpha \cdot z'' \neq \la\ra$ and $w, z \cdot \alpha, z'' \models \Psi$.
      By induction, $w, z \cdot \alpha, z' \models \Psi$ iff $w, z \cdot \alpha, z'' \models \tnf(\Psi)$.
      On the other hand, by definition, $w, z, z' \models \tnf(\lnext(\Psi))$ iff $w, z, z' \models \neg \ltail \wedge \lnext(\tnft(\Psi)) \wedge \lfin \ltail$ iff $z' \neq \la\ra$ and $w, z, z' \models \lnext(\tnf(\Psi))$.
      Hence, the claim holds.
    \item Let $\Phi$ be $\lwnext(\Psi)$ and so $\tnf(\Psi) = (\ltail \vee \lnext(\tnft(\Psi))) \wedge \lfin \ltail$.
      If $z' = \la\ra$, then $w, z, z' \models \ltail$ and so $w, z, z' \models \tnf(\Psi)$.
      Otherwise, $z' \neq \la\ra$ and so $w, z, z' \models \lwnext \Psi$ iff $w, z \cdot \alpha, z'' \models \Psi$ for $z' = \alpha \cdot z''$.
      On the other hand, $w, z \cdot \alpha, z'' \models \tnf(\Psi)$ iff $w, z, z' \models \lnext(\tnft(\Psi)) \wedge \lfin \ltail$.
      By induction, $w, z \cdot \alpha, z'' \models \tnf(\Psi)$ iff $w, z \cdot \alpha, z'' \models \Psi$.
      With $\tnf(\Psi) = \tnft(\Psi) \wedge \lfin \ltail$, the claim holds.
    \item Let $\Phi$ be $\Psi_1 \luntil \Psi_2$ and so $\tnf(\Phi) = (\neg \ltail \wedge \tnft(\Psi_1)) \luntil \tnft(\Psi_2) \wedge \lfin \ltail$.
      \\
      \textbf{$\Rightarrow$:}
      Suppose $w, z, z' \models \Psi_1 \luntil \Psi_2$.
      Then there is some $k \leq |z'|$ such that $w, z \cdot z'[..k], z'[k+1..] \models \Psi_2$ and for all $0 \leq i < k$, $w, z \cdot z'[..i], z'[i+1..] \models \Psi_1$.
      By induction, it follows that $w, z \cdot z'[..k], z'[k+1] \models \tnf(\Phi_2)$, which holds iff $w, z[..k], z'[k+1..] \models \tnft(\Phi_2) \wedge \lfin \ltail$.
      Furthermore, for every $i < k$, $w, z \cdot z'[..i], z'[i+1..] \models \Phi_1 \wedge \neg \ltail$ and so by induction $w, z[..i], z'[i+1..] \models \tnf(\Phi_1) \wedge \neg \ltail$, which implies $w, z[..i], z'[i+1..] \models \neg \ltail \wedge \tnft(\Phi_1)$.
      Hence, $w, z, z' \models (\neg \ltail \wedge \tnft(\Psi_1)) \luntil \tnft(\Psi_2) \wedge \lfin \ltail$.
      \\
      \textbf{$\Leftarrow$:}
      Suppose $w, z, z' \models (\neg \ltail \wedge \tnft(\Psi_1)) \luntil \tnft(\Psi_2) \wedge \lfin \ltail$.
      Hence, there is a $k$ such that $w, z \cdot z'[..k], z'[k+1..] \models \tnft(\Psi_2)$ and for all $0 \leq i < k$, $w, z \cdot z'[..i], z'[i+1..] \models \tnft(\Psi_1) \wedge \neg \ltail$.
      By induction, $w, z \cdot z'[..k], z'[k+1..] \models \Psi_2$ and for all $0 \leq i < k$, $w, z \cdot z'[..i], z'[i+1..] \models \Psi_1$.
      Therefore, $w, z, z' \models \Psi_1 \luntil \Psi_2$.
    \item Let $\Phi = \Psi_1 \lrelease \Psi_2$ and so $\tnf(\Phi) = (\ltail \vee \tnft(\Psi_1)) \lduntil \tnft(\Psi_2) \wedge \lfin \ltail$.
      \\
      \textbf{$\Rightarrow$:}
      Suppose $w, z, z' \models \Phi$.
      We have two cases: First, $w, z \cdot z'[..i], z'[i+1..] \models \Psi_2$ for all $i \leq |z'|$.
      By induction, for each $i$, $w, z \cdot z'[..i], z'[i+1..] \models \tnf(\Psi_2)$ and so $w, z \cdot z'[..i], z'[i+1..] \models \tnft(\Psi_2) \wedge \lfin \ltail$ and hence $w, z, z' \models (\ltail \vee \tnft(\Psi_1)) \lduntil \tnft(\Psi_2)$.
      Second, there is an $i$ such that $w, z \cdot z'[..i], z'[i+1..] \models \Psi_1$ and $w, z \cdot z'[..j], z'[j+1..] \models \Psi_2$ for all $j \leq i$.
      Again by induction, for this $i$, $w, z \cdot z'[..i], z'[i+1..] \models \tnf(\Psi_1)$ and $w, z \cdot z'[..j], z'[j+1..] \models \tnf(\Psi_2)$ for each $j \leq i$.
      Therefore, $w, z, z' \models (\ltail \vee \tnft(\Psi_1)) \lduntil \tnft(\Psi_2) \wedge \lfin \ltail$.
      \\
      {$\Leftarrow$:}
      Suppose $w, z, z' \models (\ltail \vee \tnft(\Psi_1)) \lduntil \tnft(\Psi_2) \wedge \lfin \ltail$.
      Then there is some $k \leq |z'|$ such that $w, z \cdot z'[..k], z'[k+1..] \models \ltail \vee \tnft(\Psi_1)$ and for all $0 \leq i \leq k$, $w, z \cdot z'[..i], z'[i+1..] \models \tnft(\Psi_2)$.
      Hence, by induction, $w, z \cdot z'[..k], z'[k+1..] \models \Psi_1$ and for all $0 \leq i \leq k$, $w, z \cdot z'[..i], z'[i+1..] \models \Psi_2$.
      Thus, $w, z, z' \models \Psi_1 \lrelease \Psi_2$.
      \qedhere
  \end{itemize}
\end{proofE}

In the following, each \ac{LTLf} formula is assumed to be in \ac{TNF} and we may omit the common part $\lfin \ltail$.

We continue by interpreting temporal formulas as propositional formulas by treating sub-formulas with a temporal operator as outermost connective as if they were propositional atoms.
  For a temporal formula $\Phi$, we define the set of \emph{propositional atoms} $\propatoms(\Phi)$ of $\Phi$ inductively:
  \begin{enuminline}
    \item $\propatoms(\Phi) = \{ \Phi \}$ if $\Phi$ is an atom, $\lnext$, $\luntil$, or $\lduntil$ formula
    \item $\propatoms(\Phi) = \propatoms(\Psi)$ if $\Phi = \neg \Psi$
    \item $\propatoms(\Phi) = \propatoms(\Phi_1) \cup \propatoms(\Phi_2)$ if $\Phi = \Phi_1 \wedge \Phi_2$ or $\Phi = \Phi_1 \vee \Phi_2$
  \end{enuminline}
  For a temporal formula $\Phi$, let $\Phi^p$ be $\Phi$ understood as a propositional formula over $\propatoms(\Phi)$.
  A propositional assignment $P$ of $\Phi^p$ is a partial function $P: \propatoms(\Phi) \rightarrow \{ 0, 1 \}$ that assigns truth values to the propositional atoms $\propatoms(\Phi)$. 
  We write $P \models \Phi^p$ if $P$ satisfies $\Phi^p$.
A propositional assignment $P$ can also be understood as a set of literals $\{ p \in \propatoms(\Phi) \mid P(p) = 1 \} \cup \{ \neg p \in \propatoms(\Phi) \mid P(p) = 0 \}$ and we use $P$ to denote both interchangeably.

If $\Phi$ is satisfiable, then there exists a corresponding propositional assignment:
\begin{lemmaE}\label{lma:propositional}
  Let $w$ be a world, $\Phi$ \iac{LTLf} formula, and $z$ and $z'$ traces.
  Then $w, z, z' \models \Phi$ implies there exists a propositional assignment $P$ with $P \models \Phi^p$ and $w, z, z' \models \bigwedge P$.
\end{lemmaE}
\begin{proofE}~
  [Adapted from \cite{liSATbasedExplicitLTLf2020}, Theorem 2]
  \\
  By structural induction on $\Phi$.
  \begin{itemize}
    \item If $\Phi$ is a literal, $\lnext$, $\luntil$, or $\lrelease$ formula, then $P = \{ \Phi \}$ is a satisfying propositional assignment and $w, z, z' \models \bigwedge P$.
    \item For $\Phi = \Psi_1 \wedge \Psi_2$, by induction, there $P_1$ and $P_2$ with $P_1 \models \Psi_1^p$ and $P_2 \models \Psi_2^p$.
      Let $P = P_1 \cup P_2$ be a consistent propositional assignment, in which no literal occurs both positively and negatively.
      Such a propositional assignment must exist because otherwise, $w, z, z' \not\models \Phi$.
      Then $P \models \Phi^p$ and $w, z, z' \models \bigwedge P$.
    \item For $\Phi = \Psi_1 \vee \Psi_2$, we have $w, z, z' \models \Psi_1$ or $w, z, z' \models \Psi_2$.
      Wlog, $w, z, z' \models \Psi_1$ and so by induction, there exists a propositional assignment $P_1$ with $P_1 \models \Psi_1^p$ and $w, z, z' \models \bigwedge P_1$.
      \qedhere
  \end{itemize}
\end{proofE}
The converse is not necessarily true: Let $\Phi = \lnext(a) \wedge \lnext (\neg a)$.
Clearly, $\Phi$ is not satisfiable, but $\{ \lnext(a), \lnext(\neg a) \}$ is a satisfying propositional assignment of $\Phi^p$.

We now define \ac{XNF}, where each \luntil and \lduntil operator is pushed inwards such that the only outermost temporal connective is $\lnext$:
\begin{definition}
  Let $\Phi$ be a temporal formula.  
  Its \acf{XNF} $\xnf(\Phi)$ is defined recursively as follows:
\begin{enumerate}
  \item $\xnf(\Phi) = \Phi$ if $\Phi$ is $\top, \bot$, a \ctwo-fluent sentence, or $\lnext \Psi$;
  \item $\xnf(\Phi_1 \wedge \Phi_2) = \xnf(\Phi_1) \wedge \xnf(\Phi_2)$;
  \item $\xnf(\Phi_1 \vee \Phi_2) = \xnf(\Phi_1) \vee \xnf(\Phi_2)$;
  \item $\xnf(\Phi_1 \luntil \Phi_2) = \xnf(\Phi_2) \vee (\xnf(\Phi_1) \wedge \lnext(\Phi_1 \luntil \Phi_2))$;
  \item $\xnf(\Phi_1 \lrelease \Phi_2) = \xnf(\Phi_2) \wedge (\xnf(\Phi_1) \vee \lnext(\Phi_1 \lrelease \Phi_2))$.
\end{enumerate}
\end{definition}
It can be shown that $\Phi$ and $\xnf(\Phi)$ are equivalent:
\begin{theoremE}
  Let $\Phi$ be a temporal formula, $w$ a world, and $z$ and $z'$ finite traces.
  Then $w, z, z' \models \Phi$ iff $w, z, z' \models \xnf(\Phi)$.
\end{theoremE}
\begin{proofE}
  By structural induction on $\Phi$.
  \begin{itemize}
    \item If $\Phi$ is $\top, \bot$, a \ctwo-fluent sentence, or $\lnext \Psi$, then $\xnf(\Phi) = \Phi$ and the claim holds.
    \item The Boolean cases follow immediately by induction.
    \item Let $\Phi = \Psi_1 \luntil \Psi_2$.
      By semantics of \luntil, $w, z, z' \models \Phi$ iff $w, z, z' \models \Psi_2$ or $w, z, z' \models \Psi_1 \wedge \lnext(\Psi_1 \luntil \Psi_2)$.
      By induction, $w, z, z' \models \Psi_2$ iff $w, z, z' \models \xnf(\Psi_2)$ and $w, z, z' \models \Psi_1 \wedge \lnext(\Psi_1 \luntil \Psi_2)$ iff $w, z, z' \models \xnf(\Psi_1) \wedge \lnext(\xnf(\Psi_1 \luntil \Psi_2))$ and so the claim follows.
    \item Let $\Phi = \Psi_1 \lduntil \Psi_2$.
      By semantics of \lduntil, $w, z, z' \models \Phi$ iff $w, z, z' \models \Psi_2$ and $w, z, z' \models \Psi_1 \vee \lnext(\Psi_1 \lduntil \Psi_2)$.
      By induction, $w, z, z' \models \Psi_2$ iff $w, z, z' \models \xnf(\Psi_2)$ and $w, z, z' \models \Psi_1 \vee \lnext(\Psi_1 \lduntil \Psi_2)$ iff $w, z, z' \models \xnf(\Psi_1) \vee \lnext(\xnf(\Psi_1 \lduntil \Psi_2))$ and so the claim follows.
      \qedhere
  \end{itemize}
\end{proofE}


For a propositional assignment $P$ of $\Phi^p$ in \ac{XNF}, we define 
\begin{iteminline}
  \item $L(P) = \{ l \mid l \in P \text{ is a literal other than $(\neg) \ltail$ } \}$
  \item $X(P) = \{ \theta \mid \lnext \theta \in P \}$
  \item $T(P) = \top$ if $\ltail \in P$ and $T(P) = \bot$ otherwise
\end{iteminline}

\ac{XNF} allows us to track the partial satisfaction of a temporal formula over a trace.
After each action, we will determine each satisfying assignment $P$ such that $L(P)$ is satisfied by the current state and we will track $X(P)$ in the remaining trace.
We will use this in the following to construct a game arena that tracks the satisfaction of a temporal formula $\Phi$.

\section{Approach}
\label{sec:approach}

Our goal is to determine an execution of a given \golog program that satisfies the given temporal formula, for all possible environment behaviors.
The controller must determine which actions to execute; more specifically, which branch to follow in all nondeterministic choices of the program, while not restricting the environment in its actions.
Formally, our goal is to find a successful policy, defined as follows:
\begin{definition}[Policy]\label{def:policy}
  Let $\prog = (\bat, \delta)$ be a \golog program and $\actions = \actions_C \dot{\cup} \actions_E$ a partition of the actions \actions of \prog into controllable and environment actions.
  A \emph{policy} is a partial mapping $\pi: \mathcal{W} \times \mathcal{Z} \times \sub(\delta) \rightarrow 2^{\mathcal{A}}$ such that:
    \begin{enumerate}
      \item \label{def:policy:init} if $w \models \bat$, then $\pi$ is defined on $(w, \la\ra, \delta)$;
      \item \label{def:policy:applicable} if $\alpha \in \pi(w, z, \rho)$, then $\la z, \rho \ra \xrightarrow{w} \la z \cdot \alpha, \rho' \ra$ for some $\rho' \in \sub(\delta)$;
      \item \label{def:policy:defined-on-succ} if $\alpha \in \pi(w, z, \rho)$ and $\la z, \rho \ra \xrightarrow{w} \la z \cdot \alpha, \rho' \ra$, then $\pi$ is defined on $(w, z \cdot \alpha, \rho')$;
      \item \label{def:policy:env-actions} if $\alpha \in \actions_E$ and $\la z, \rho \ra \xrightarrow{w} \la z \cdot \alpha, \rho' \ra$ for some $\rho' \in \sub(\delta)$, then $\alpha \in \pi(w, z, \rho)$;
      \item \label{def:policy:non-blocking} if $\pi(w, z, \rho) = \emptyset$, then $\la z, \rho \ra \in \final(w)$.
    \end{enumerate}
\end{definition}

Intuitively, a policy chooses a subset $\pi(w, z, \rho)$ from all possible actions in the current configuration $\la z, \rho \ra$ and world $w$.
From this subset, the environment then chooses one action to be executed.
The agent's choices are restricted: Every possible environment action must be selected, hence the agent can never limit the environment's choices.

A policy $\pi$ induces a set of traces $\|\pi\|_w$ in world $w$, where $z = \la \alpha_1, \ldots, \alpha_n \ra \in \|\pi\|_w$ if there are $\rho_1, \ldots, \rho_n$ such that
\begin{enuminline}
  \item $\la \la \ra, \delta \ra \xrightarrow{w} \la z[..1], \rho_1 \ra \xrightarrow{w} \cdots \xrightarrow{w} \la z, \rho_n \ra$
  \item $\alpha_{i+1} \in \pi(w, z[..i], \rho_i)$
  \item $\pi(w, z, \rho_n) \subseteq \actions_E$ and $\la z, \rho_n \ra \in \final(w)$
\end{enuminline}
Hence, the environment may choose to terminate the execution if $\la z, \rho \ra$ is a final configuration and the agent chose no further actions to execute.
Note that by definition, a policy is a restriction of the program execution, i.e., $\|\pi\|_w \subseteq \|\delta\|_w$.
We call a policy \emph{terminating} if for every infinite sequence of $\pi$-compatible configurations $\la \la\ra, \delta \ra, \la z_1, \rho_1 \ra, \la z_2, \rho_2 \ra, \ldots$
and for every $i$, there is a $j \geq i$ such that $\pi(w, z_j, \rho_j) \subseteq \actions_E$ and $\la z_j, \rho_j \ra \in \final(w)$.
Intuitively, a terminating policy ensures that at any point of the execution trace, there is some future final configuration where the policy does not choose any agent actions and hence the environment may terminate.
A policy may still result in an infinite trace if the environment continues to select actions indefinitely.
However, we exclude those from consideration as we assume that the environment eventually stops.
%
%
%
We can now formalize our goal:
\begin{definition}[Synthesis Problem]
  Given a \golog program $\prog = (\bat, \delta)$ and a temporal formula $\Phi$, find a \emph{terminating policy} $\pi$ for $\prog$ that \emph{satisfies} $\Phi$,
  i.e., for every world $w$ with $w \models \bat$ and every $z \in \|\pi\|_w$, it holds that $w, \la\ra, z \models \Phi$.
\end{definition}
We note that it is in general undecidable to determine whether a satisfying policy exists.
In \cite{zarriessDecidabilityVerifyingLTL2014, zarriessDecidableVerificationGolog2016} it was shown that the related verification problem (a special case of the synthesis problem) becomes decidable if (1) \ctwo is used as base logic, (2) successor state axioms are acyclic, and (3) ``pick operators'' are disallowed, i.e., all actions in the program are ground.
Furthermore, dropping any of these three restrictions while maintaining the other two immediately leads to undecidability: for (1) this is due to the undecidability of FOL, and for (2) and (3) due to the possibility of reducing the halting problem for Turing machines to the verification problem.

In the following, applying the same three restrictions, we describe a sound and complete method for determining a terminating policy $\pi$ that
satisfies $\Phi$.
We will do so by constructing a finite game arena \ts that captures the possible program executions while tracking the satisfaction of $\Phi$.
Once we have constructed \ts, we can use a game-theoretic approach to determine a terminating policy that satisfies $\Phi$.
However, as both the number of worlds satisfying \bat and the number of reachable program configurations is generally infinite, we first need to construct a finite abstraction based on \emph{characteristic graphs} and \emph{types}.

\subsection{Characteristic Graphs}
\emph{Characteristic graphs}~\cite{classenLogicNonterminatingGolog2008} provide a finite encoding of the reachable program configurations.
In such a graph, the nodes correspond to programs $\rho$, intuitively representing what remains to be executed, while an edge $\rho \progtrans{\alpha}{\psi} \rho'$ encodes that a transition is possible from $\rho$ to $\rho'$ through action $\alpha$, if formula $\psi$ holds.
In addition, each program $\rho$ has an associated {\em termination condition} $\termcond{\rho}$, in the form of a fluent formula.

\begin{definition}[Characteristic Graph]
  Given a program expression $\delta$, the {\em termination condition $\termcond{\delta}$ of $\delta$} is a fluent formula inductively defined as follows:
  \begin{sizeddisplay}
    \begin{align*}
      \termcond{\alpha} &= \false \text{ if $\alpha$ is a ground action }
                        &\termcond{\phi?} &= \phi
      \\
      \termcond{\delta_1\seq\delta_2} &= \termcond{\delta_1} \land \termcond{\delta_2}
      &\termcond{\delta_1\nondet\delta_2} &= \termcond{\delta_1} \lor \termcond{\delta_2}
      \\
      \termcond{\delta_1\conc\delta_2} &= \termcond{\delta_1} \land \termcond{\delta_2}
      &\termcond{\iter{\delta}} &= \true
    \end{align*}
  \end{sizeddisplay}
  For any program expression $\delta$, the set of {\em outgoing edges $\delta \progtrans{\alpha}{\psi} \rho$} with action $\alpha$ and guard condition $\psi$ to resulting program $\rho$ is defined inductively as follows:
  \begin{itemize}
  \item $\alpha \progtrans{\alpha}{\true} \nil$, if $\alpha$ is a primitive action;
  \item $(\delta_1\seq\delta_2) \progtrans{\alpha}{\psi} (\rho\seq\delta_2)$, if $\delta_1 \progtrans{\alpha}{\psi} \rho$;
  \item $(\delta_1\seq\delta_2) \progtrans{\alpha}{\termcond{\delta_1} \land \psi} \rho$, if $\delta_2 \progtrans{\alpha}{\psi} \rho$;
  \item $(\delta_1\nondet\delta_2) \progtrans{\alpha}{\psi} \rho$, if $\delta_1 \progtrans{\alpha}{\psi} \rho$ or $\delta_2 \progtrans{\alpha}{\psi} \rho$;
  \item $(\delta_1\conc\delta_2) \progtrans{\alpha}{\psi} (\rho\conc\delta_2)$, if $\delta_1 \progtrans{\alpha}{\psi} \rho$;
  \item $(\delta_1\conc\delta_2) \progtrans{\alpha}{\psi} (\delta_1\conc\rho)$, if $\delta_2 \progtrans{\alpha}{\psi} \rho$;
  \item $\iter{\delta} \progtrans{\alpha}{\psi} (\rho\seq\iter{\delta})$, if $\delta \progtrans{\alpha}{\psi} \rho$.
  \end{itemize}
  For any program expression $\delta$, the corresponding {\em characteristic graph} is given by $\chargraph = \langle v_0,V,E \rangle$, where $v_0 = \delta$ (initial node), and the nodes $V$ and edges $E$ are the smallest sets such that
  \begin{enuminline}
  \item $\delta \in V$
  \item if $\delta' \in V$ and $\delta' \progtrans{\alpha}{\psi} \delta''$, then $\delta'' \in V$ and $\delta' \progtrans{\alpha}{\psi} \delta'' \in E$
  \end{enuminline}
\end{definition}

We denote the set $V$ with $\sub(\delta)$, the {\em subprograms reachable from $\delta$}.
We note~\cite{classenLogicNonterminatingGolog2008}:
\begin{lemmaE}
  \label{lem:CGraphsCorrectness}
  For any program $\delta$, \chargraph is finite, and for any world $w$, situation $z$, and $\delta' \in\sub(\delta)$, it holds that
  \begin{enuminline}
  \item $\la z, \delta' \ra \in\final(w)$ iff $w,z \models \termcond{\delta'}$
  \item $\la z, \delta' \ra \xrightarrow{w} \la z \cdot \alpha, \delta'' \ra$ iff $\delta'\progtrans{\alpha}{\psi}\delta''$ and $w,z \models \psi$
  \end{enuminline}
\end{lemmaE}
Characteristic graphs therefore exactly capture the program transition semantics.
We can hence use them as finite abstractions of the reachable program configurations.
Also, using characteristic graphs, there is a (simple to test) sufficient condition for programs being situation-determined:
\begin{lemmaE}
If every ground action $\alpha$ occurs at most once among the
outgoing edges of every node in $\chargraph$, then $\delta$ is
situation-determined.
\end{lemmaE}
\begin{proofE}[][Proof Idea]
By induction on the length of traces starting in $\la z,\delta \ra$,
using Lemma \ref{lem:CGraphsCorrectness}.
\end{proofE}

\subsection{Types}

With characteristic graphs, we already have a finite representation of the possible program configurations.
However, there are additional sources of infiniteness.
For one, during the execution of a program, we may accumulate infinitely many effects.
Second, there are infinitely many possible worlds that satisfy the \ac{BAT} \bat.
However, for acyclic \acp{BAT}, it has been shown that the set of possible effects is finite, and that the set of worlds that satisfy \bat can be represented by a finite set of equivalence classes, so-called \emph{types of worlds}~\cite{zarriessDecidableVerificationGolog2016}.
We will now describe how to construct types for a given \ac{BAT} \bat.

As our programs may only mention finitely many ground actions, we can rewrite the \acp{SSA} of an acyclic \ac{BAT} by grounding the effects.
This is done by replacing each \ac{SSA} for a fluent $F(\vec{x})$ by a set of instantiated formulas, one for each $\alpha \in \actions$, of the form
$\square [\alpha] F(\vec{x}) \equiv \gammap \vee F(\vec{x}) \wedge \neg \gamman$.
As each $\gamma_F^\pm$ is a disjunction of formulas of the form $\exists \vec{y}.(a = A(\vec{v}) \wedge \epsilon \wedge \kappa)$, the resulting positive effect condition \gammap  is equivalent to a disjunction of the form
$\epsilon_1 \wedge \kappa_1 \vee \ldots \vee \epsilon_n \wedge \kappa_n$,
which allows us to write \gammap  as a set of pairs $\gammap = \bigvee_i \{ ( \epsilon_i, \kappa_i ) \}_i$.
We write $(\epsilon, \kappa) \in \gammap$ if $(\epsilon, \kappa)$ occurs in the disjunction (analogously for \gamman).
For a fluent $F$, the set of positive effect descriptors is then defined as
$\effpos{F} := \{ \effform \mid (\effform,\effcon) \in \gammap \text{ for some } \alpha \in \Act \}$,
and similarly for negative effect descriptors $\effneg{F}$. 
Hence, we can write a set of effects $E$ as a set of pairs $E = \{ \la F^\pm_i, \effform_i \ra \}_i$, where $\effform_i \in \effpos{F}$ or $\effform_i \in \effneg{F}$.
We define a variant of \emph{regression} on such a set of effects:
\begin{definition}[Regression]
  Let $E$ be a set of effects and $\varphi$ a $C^2$ fluent formula.
  The \emph{regression of $\varphi$ through $E$}, denoted by $\reg[E, \varphi]$ is a $C^2$ fluent formula obtained from $\varphi$ by replacing each occurrence of a fluent $F(\vec{v})$ in $\varphi$ by the formula
  $
  F(\vec{v}) \wedge \bigwedge_{\la F^-, \effform \ra \in E} \neg \effform^{\vec{x}}_{\vec{v}} \vee \bigvee_{\la F^+, \effform \ra \in E} \effform^{\vec{x}}_{\vec{v}}
  $.
\end{definition}

Furthermore, in an acyclic \ac{BAT}, the effect descriptor $\effform$ of a fluent $F$ with $\fd(F) = i$ may only mention fluents with depth strictly smaller than $i$.
Thus, when regressing the effect descriptor $\effform$ of a fluent $F$ with $\fd(F) = i$, only effects on fluents with depth strictly smaller than $i$ are relevant.
Hence, for a \golog program $\prog = (\bat, \delta)$ with an acyclic \ac{BAT} \bat, there are only finitely many possible effects that can be generated by action sequences from \actions.
We denote the \emph{set of all relevant effects} on all fluents with depth $\leq j$ with $j = 0,\ldots,\fdbat$ by $\effset{j}$, and define it as follows: 
\begin{sizeddisplay}
\begin{align*}
	\begin{aligned}
    \effset{0} \eqdef {}& \{ \adddeleff{F}{\effform} \mid \fdfluent{F} = 0, \effform \in \effneg{F} \cup \effpos{F} \} \\
    \effset{i} \eqdef {}& \effset{i-1} \\ & \cup \{ \deleff{F}{\Regress{\E}{\effform}}
		\mid
		\fdfluent{F} = i, \effform \in \effneg{F}, \E \subseteq \effset{i-1} \}
    \\
		& \cup\{ \addeff{F}{ \Xi  }
		\mid
		\begin{aligned}[t]
			& \fdfluent{F} = i, \phi \in \effpos{F}, \E \subseteq \effset{i-1},  \\
			& X \subseteq \effneg{F} \times 2^{\effset{i-1}} \}
		\end{aligned} \\
		& \text{with } \Xi \eqdef \bigl( \Regress{\E}{\phi} \land \bigwedge\limits_{ (\effform, \E') \in X} \neg \Regress{\E'}{\effform}  \bigr)
    \\
      \effsetcomplete \eqdef {}& \effset{n} \text{ with $\fdbat = n$ }
	\end{aligned}
\end{align*}
\end{sizeddisplay}

Additionally, we define the \emph{context of a program} $\context(\prog)$ as the set of relevant \ctwo-fluent sentences that occur in the initial theory, in context conditions of the instantiated \acp{SSA}, in guards and termination conditions of the characteristic graph, and in the temporal formula, and we ensure that the context is closed under negation.
We can now define \emph{types}:
\begin{definition}[Type of a world]
  Let $\prog = (\bat, \delta)$ be a \golog program with an acyclic \ac{BAT} $\bat = \bat_0 \cup \bat_\text{post}$ w.r.t.\ a finite set of ground actions \actions.
  Furthermore, let $\context(\prog)$ be the context of \prog and \effects the set of all relevant effects.
  The \emph{set of all type elements} is given by
  $\typeelements(\prog) \eqdef \{ (\psi, E) \mid \psi \in \context(\prog), E \subseteq \effects \}$.
  A \emph{type w.r.t.\ \prog} is a set $\tau \subseteq \typeelements(\prog)$ that satisfies:
  \begin{enumerate}
    \item For all $\psi \in \context(\prog)$ and all $E \subseteq \effects$ it holds that either $(\psi, E) \in \tau$ or $(\neg \psi, E) \in \tau$;
    \item There exists a world $w \in \worlds$ such that
      $w \models \bat_0 \cup \{ \reg[E, \psi] \mid (\psi, E) \in \tau \}$.
  \end{enumerate}
  The \emph{set of all types w.r.t.\ \prog} is denoted by $\types(\prog)$.
  The \emph{type of a world $w \in \worlds$ w.r.t.\ \prog} is given by
  $\type(w) \eqdef \{ (\psi, E) \in \typeelements(\prog) \mid w \models \reg[E, \psi] \}$.
\end{definition}

\begin{definition}
  Let $\tau \in \types(\prog)$, $E \subseteq \effects$, and $\alpha \in \actions$.
  The \emph{effects of executing $\alpha$ in $(\tau, E)$} are given by
  \begin{sizeddisplay}
  \begin{align*}
    \mathcal{E}_\mathcal{D}(\tau, E, \alpha) \eqdef{} &\{ \la F^+, \effform \ra \mid \exists (\effform, \effcon) \in \gammap \text{ s.t. } (\effcon, E) \in \tau \} ~\cup
      \\
      &\{ \la F^-, \effform \ra \mid \exists (\effform, \effcon) \in \gamman \text{ s.t. } (\effcon, E) \in \tau \}
  \end{align*}
  \end{sizeddisplay}
\end{definition}
  
\begin{definition}
  Let $\varphi$ be a $C^2$ fluent formula and $E_0$ and $E_1$ two sets of effects.
  The accumulation $E_0 \triangleright E_1$ of $E_0$ and $E_1$ is defined as follows:
  \begin{sizeddisplay}
  \begin{align*}
    &E_0 \triangleright E_1 \eqdef 
    \{ \la F^\pm, \reg[E_0, \varphi] \ra \mid \la F^\pm, \varphi \ra \in E_1 \}
    \\
    &\cup \{ \la F^+, ( \varphi \wedge \bigwedge_{\mathclap{\la F^-, \varphi \ra \in E_1}} \neg \reg[E_0, \varphi'] ) \ra \mid \la F^+, \varphi \ra \in E_0 \}
                                  \cup \{ \la F^-, \varphi \ra \in E_0 \}
  \end{align*}
  \end{sizeddisplay}
\end{definition}

Let $w$ be a world with $w \models \bat$, $\type(w) = \tau$, and $z = \la \alpha_1, \ldots, \alpha_n \ra$ a trace.
We define
  $E_0 \eqdef \emptyset$
  and
  $E_i \eqdef E_{i-1} \triangleright \mathcal{E}_\bat(\tau, E, \alpha)$  for $1 \leq i \leq n$.
%
We also write $E_z$ for the effect $E_n$ that is generated by executing $z = \la \alpha_1, \ldots, \alpha_n \ra$ in $w$.
The following theorem shows the correctness of the construction~\cite{zarriessDecidableVerificationGolog2016}:\todo{This is very similar to the 2016 paper, but the theorem in this form is new}
  \todo{Add proof?}
\begin{theorem}\label{thm:effectAccumulation}
  Let $\prog = (\bat, \delta)$ be a \golog program, $w$ a world with $w \models \bat$, and $z \in \actions^*$ a trace. Then
  $w, z \models \phi$ iff $(\phi, E_z) \in \type(w)$.
\end{theorem}

Hence, types provide a finite representation of the worlds satisfying \bat and all effects that can be generated by $\delta$.

\subsection{Game Arena}

With types, characteristic graphs, and \ac{XNF} formulas, we can define a game arena \ts that captures the possible executions of a program \prog while tracking the satisfaction of $\Phi$:
\begin{definition}
  \label{def:TransitionSystem}
  Let $\prog = (\bat, \delta)$ be a \golog program and $\Phi$ a temporal formula.
  The game arena $\ts = (\tsstates, \tsstates_0, \operatorname{\rightarrow}, \tsstates_F, \tsstates_A)$ for $\prog$ and $\Phi$ is defined as follows:
\begin{itemize}
  \item Each state $s \in \tsstates$ is of the form $s = (\tau, E, A, \rho)$ where
    \begin{enumerate}
      \item $\tau \in \types(\mathcal{G})$;
      \item $\rho \in \sub(\delta)$ is a node of the characteristic graph;
      \item $E \subseteq \effects$;
      \item $A = \{ (\chi_i, \theta_i) \}_i$, where
        \begin{enuminline}[label=,itemjoin*={,\ }]
          \item $\chi_i \subseteq \cl(\Phi)$
          \item $\theta_i \in \{ \top, \bot \}$
        \end{enuminline}
    \end{enumerate}
  \item A state $s = (\tau, E, A, \rho)$ is an initial state $s \in \tsstates_0$ if
    \begin{enumerate}
      \item $\tau = \type(w)$ for some $w$ with $w \models \bat$;
      \item $\rho = \delta$ is the initial program expression;
      \item $E = \emptyset$;
      \item $(\chi, \theta) \in A$ iff there is a propositional assignment $P$ of $\xnf(\Phi)^p$ such that
        \begin{iteminline}
          \item $\{ (\psi, E) \mid \psi \in L(P) \} \subseteq \tau$
          \item $\chi = X(P)$
          \item $\theta = T(P)$
        \end{iteminline}
    \end{enumerate}
  \item There is a transition $s_1 \xrightarrow{\alpha} s_2$ from  $s_1 = (\tau, E_1, A_1, \rho_1)$ to $s_2 = (\tau, E_2, A_2, \rho_2)$ if
    \begin{enumerate}
      \item there is an edge $\rho_1 \progtrans{\alpha}{\psi} \rho_2$ in \chargraph with $(\psi, E_1) \in \tau$;
      \item $E_2 = E_1 \triangleright \mathcal{E}_\mathcal{D}(\tau, E_1, \alpha)$;
      \item $(\chi_2, \theta_2) \in A_2$ if there is a propositional assignment $P$ of $\xnf(\bigwedge \chi_1^p)$ for some $(\chi_1, \theta_1) \in A_1$ such that
        \begin{enuminline}[itemjoin={,\ },itemjoin*={, and\ }]
         \item $\theta_1 = \bot$
          \item $\{ (\psi, E_2) \mid \psi \in L(P) \} \subseteq \tau$
          \item $\chi_2 = X(P)$
          \item $\theta_2 = T(P)$
        \end{enuminline}
    \end{enumerate}
\end{itemize}
A state $s = (\tau, E, A, \rho)$ is \emph{final} if $(\varphi(\rho), E) \in \tau$ and \emph{accepting} if $(\emptyset, \top) \in A$.
We denote the set of all final states with $\tsstates_F$ and the set of all accepting states with $\tsstates_A$.
We also write $\type(s) = \tau$ for the type of the world in $s$.
\end{definition}

Each state consists of
\begin{enuminline}
  \item a type $\tau$, representing an equivalence class of worlds
  \item a node $\rho$ from the characteristic graph, capturing the remaining program and its termination condition
  \item  a set of accumulated effects $E$
  \item a set of temporal formulas $A$, which must be satisfied in the remaining execution to fulfill the specification $\Phi$
\end{enuminline}
The initial states are those states with the initial program expression and no accumulated effects.
Furthermore, regarding the temporal formula $\Phi$ and $A$ of an initial state, we first compute all the propositional assignments of $\xnf(\Phi)^p$.
For each assignment $P$, we check whether the local part $L(P)$ is satisfied by the state.
If so, the pair $(\chi, \theta) = (X(P), T(P))$ is added to $A$, which intuitively states that $\chi$ must be satisfied in the future and the program should terminate if $\theta$ is true.
For transitions, we first check whether there is an edge in the characteristic graph that allows the execution of the next action.
If so, we accumulate the effects and check whether there is a propositional assignment of $\xnf(\bigwedge \chi_1^p)$ for some $(\chi_1, \theta_1) \in A_1$ that allows the satisfaction of the temporal formulas in $A_2$.
Similar to the initial states, we do so by checking whether the local part $L(P)$ is satisfied by the current state and tracking $X(P)$ and $T(P)$ in the future.

By definition, a state is \emph{final} if the program may terminate and it is \emph{accepting} if $\Phi$ is satisfied.
Also note that \ts is finite as both types and reachable sub-programs are finite.
It is also deterministic, as \prog is situation-determined and for action successors, the satisfying assignments of $\xnf(\Phi)^p$ are collected in a single successor state.

\textEnd{We can show that \ts indeed tracks the program executions of \prog:}
\begin{theoremEnd}[all end]{lemma}
  \label{lma:transitionSystemProgramTraces}
  Let $z = \la \alpha_1, \ldots, \alpha_n \ra \in \traces$ be an arbitrary trace and $\prog = (\bat, \delta)$ a \golog program.
  Then $z \in \| \delta \|_w$ for some world $w$ with $w \models \bat$ iff there is a path $s_0 \xrightarrow{\alpha_1} s_1 \xrightarrow{\alpha_2} \cdots \xrightarrow{\alpha_n} s_n$ in \ts such that $s_0$ is an initial state with $\type(s_0) = \type(w)$ and $s_n$ is a final state.
\end{theoremEnd}
\begin{proofE}
  We first show by induction on $n$ that $\la \la\ra, \delta \ra \xrightarrow{w} \la z[..1], \delta_1 \ra \xrightarrow{w} \cdots \xrightarrow{w} \la z, \rho_n \ra$ in \ts iff $s_0 \xrightarrow{\alpha_1} s_1 \xrightarrow{\alpha_2} \cdots \xrightarrow{\alpha_n} s_n$ in \ts such that $\type(s_0) = \type(w) = \tau$ and where for every context formula $\phi \in \context(\prog)$, we have $w, z[..i] \models \phi$ iff $(\phi, E_i) \in \type(s_i)$.
  \\
  Let each $s_i$ be of the form $s_i = (\tau, E_i, A_i, \rho_i)$.
  \\
  \textbf{Base case.} $n = 0$:
  By definition of \ts, if $w \models \bat$, then there is an initial state $s_0$ with $\type(s_0) = \type(w)$.
  Also, $E_0 = \emptyset$ and so $(\phi, E_0) \in \type(s_0)$ iff $w \models \phi$.
  \\
  \textbf{Induction step.}
  By definition, there is a transition $s_i \xrightarrow{\alpha_i} s_{i+1}$ iff $\rho_i \progtrans{\alpha_i}{\psi} \rho_{i+1}$ and $(\psi, E_i) \in \tau$.
  By induction, $(\psi, E_i) \in \tau$ iff $w, z[..i] \models \psi$ and with Lemma~\ref{lem:CGraphsCorrectness}, it follows that $s_i \xrightarrow{\alpha_i} s_{i+1}$ iff $\la z[..i], \rho_i \ra \xrightarrow{w} \la z[..i+1], \rho_{i+1} \ra$.
  By definition $E_2 = E_1 \triangleright \mathcal{E}_\mathcal{D}(\tau, E_1, \alpha_i)$ and so, with Theorem~\ref{thm:effectAccumulation}, for every $\phi \in \context(\prog)$, we have $(\phi, E_{i+1}) \in \tau$ iff $w, z[..i+1] \models \phi$.
  \smallskip
  \\
  Now, by Lemma~\ref{lem:CGraphsCorrectness}, $z \in \| \delta \|_w$ iff $w, z \models \varphi(\delta)$.
  From above, it follows that $w, z \models \varphi(\delta)$ iff $(\varphi(\delta), E_n) \in \tau$ iff $s_n$ is final.
\end{proofE}

\textEnd{Regarding the temporal formula $\Phi$, the following two lemmas show that \ts indeed tracks the satisfaction of $\Phi$:}
\begin{theoremEnd}[all end]{lemma}
  \label{lma:transitionSystemPhiSatisfied}
  Suppose $w, \la\ra, z \models \Phi$ with $z = \la \alpha_1, \ldots, \alpha_n \ra$.
  Then there is a path $s_0 \xrightarrow{\alpha_1} s_1 \xrightarrow{\alpha_2} \cdots \xrightarrow{\alpha_n} s_n$ in \ts starting in an initial state $s_0$ with $\type(s_0) = \type(w)$ such that $s_i = (\tau, E_i, A_i, \rho_i)$
  and such that for every $i \leq n$, $w, z[..i], z[i+1..] \models \bigwedge_{\Psi \in \chi_{i}} \lnext \Psi$ for some $(\chi_i, \theta_i) \in A_i$.
\end{theoremEnd}
\begin{proofE}
  By induction on $i$.
  \\
  \textbf{Base case.} Let $i = 0$.
  By Lemma~\ref{lma:propositional}, there is a propositional assignment $P_0$ of $\xnf(\Phi)^p$ with $w, \la\ra, z \models \bigwedge P_0$ and therefore also $w, \la\ra, z \models \bigwedge_{\Psi \in X(P_0)} \lnext \Psi$.
  By definition of \ts, $(X(P_0), \theta_0) \in A_0$ for some $\theta_0$.
  \\
  \textbf{Induction step.}
  By induction, $w, z[..i-1], z[i..] \models \bigwedge_{\Psi \in \chi_{i-1}} \lnext \Psi$ for some $(\chi_{i-1}, \theta_{i-1}) \in A_{i-1}$.
  Hence, $w, z[..i], z[i+1..] \models \bigwedge \chi_{i-1}$.
  By Lemma~\ref{lma:propositional}, there is a propositional assignment $P_i$ of $\xnf(\bigwedge \chi_{i-1})^p$ with $w, z[..i], z[i+1..] \models \bigwedge P_i$ and hence also $w, z[..i], z[i+1..] \models \bigwedge_{\Psi \in X(P_i)} \lnext \Psi$.
  By definition of \ts, $(X(P_i), T(P_i)) \in A_i$.
\end{proofE}

\begin{theoremEnd}[all end]{lemma}
  \label{lma:transitionSystemAccepting}
  Let $s_0 \xrightarrow{\alpha_1} s_1 \xrightarrow{\alpha_2} \cdots \xrightarrow{\alpha_n} s_n$ be a path in \ts starting in an initial state $s_0$ with $\type(s_0) = \type(w)$ and ending in an accepting state $s_n$.
  Suppose $s_i = (\tau, E_i, A_i, \rho_i)$ for each $i$.
  Then there is a sequence $\chi_0, \ldots, \chi_n$ such that for each $i$, $(\chi_i, \theta_i) \in A_i$ and $w, z[..i], z[i+1..] \models \bigwedge_{\Psi \in \chi_i} \lnext \Psi$.
\end{theoremEnd}
\begin{proofE}
  By induction on $i$ from $n$ to $0$.
  \\
  \textbf{Base case.} Let $i = n$.
  Then $s_n$ is accepting and so there is $(\chi_n, \theta_n) \in A_n$ with $\theta_n = \top$ and $\chi_n = \emptyset$.
  Trivially, $w, z, \la\ra \models \bigwedge_{\Psi \in \chi_n} \lnext \Psi$.
  \\
  \textbf{Induction step.}
  By induction, there is $(\chi_i, \theta_i) \in A_i$ such that $w, z[..i], z[i+1..] \models \bigwedge_{\Psi \in \chi_i} \lnext \Psi$.
  By definition of \ts, there is a propositional assignment $P_i$ of $\xnf(\bigwedge \chi_{i-1})^p$ (as otherwise $A_i = \emptyset$) such that $X(P_i) = \chi_i$, $T(P_i) = \theta_i$, and $\{ (\psi, E_i) \mid \psi \in L(P_i) \} \subseteq \tau$.
  Therefore, $w, z[..i], z[i+1..] \models L(P) \wedge T(P) \wedge \bigwedge_{\Psi \in \chi_i} \lnext \Psi$ and so $w, z[..i], z[i+1..] \models \xnf(\bigwedge \chi_{i-1})$.
  It directly follows that $w, z[..i-1], z[i...] \models \bigwedge_{\Psi \in \chi_{i-1}} \lnext \Psi$.
  Again by definition of \ts, $(\chi_{i-1}, \theta_{i-1}) \in A_{i-1}$.
\end{proofE}

\textEnd{Combining the results, we obtain the following theorem:}
We can show that \ts indeed corresponds to the executions of \prog while tracking the satisfaction of $\Phi$:
\begin{theoremE}
  Every execution of $\prog = (\bat, \delta)$ satisfies $\Phi$ iff every reachable final state of \ts is accepting.
\end{theoremE}
\begin{proofE}
  ~
  \\
  \textbf{$\Rightarrow$:}
  By contradiction.
  Suppose there is a reachable final state $s_n = (\tau, E, A, \rho)$ that is not accepting and let $s_0 \xrightarrow{\alpha_1} s_1 \xrightarrow{\alpha_2} \cdots \xrightarrow{\alpha_n} s_n$ be a path in \ts starting in an initial state $s_0$ with $\type(s_0) = \type(w)$ and ending in $s_n$.
  By Lemma~\ref{lma:transitionSystemProgramTraces}, $z = \la \alpha_1, \ldots, \alpha_n \ra \in \| \delta \|_w$.
  By assumption, $w, \la\ra, z \models \Phi$ and so, with Lemma~\ref{lma:transitionSystemPhiSatisfied}, $w, z, \la\ra \models \bigwedge_{\Psi \in \chi} \lnext \Psi$ for some $(\chi, \theta) \in A$.
  Clearly, $w, z, \la\ra \not\models \lnext \Psi$ for arbitrary $\Psi$, and so $\chi = \emptyset$.
  Furthermore, $w, z, \la\ra \models \ltail$ and so $\theta = \top$.
  But then, $s_n$ is accepting, a contradiction.
  \\
  \textbf{$\Leftarrow$:}
  By contradiction.
  Suppose there is a trace $z = \la \alpha_1, \ldots, \alpha_n \ra$ such that $z \in \| \delta \|_w$ but $w, \la\ra, z \not\models \Phi$.
  By Lemma~\ref{lma:transitionSystemProgramTraces}, there is a path $s_0 \xrightarrow{\alpha_1} s_1 \xrightarrow{\alpha_2} \cdots \xrightarrow{\alpha_n} s_n$ in \ts starting in an initial state $s_0$ with $\type(s_0) = \type(w)$ and ending in a final state $s_n$.
  By assumption, $s_n$ is accepting.
  By Lemma~\ref{lma:transitionSystemAccepting}, $w, \la\ra, z \models  \bigwedge_{\Psi \in \chi} \lnext \Psi$ for some $(\chi, \theta) \in A$ and $s_0 = (\tau, E, A, \rho)$.
  By definition, for each $\chi \in A$, there is a propositional assignment $P$ of $\xnf(\bigwedge \chi^p)$ such that $w, \la\ra, z \models \bigwedge L(P)$ and $\chi = X(P)$.
  But then, $w, \la\ra, z \models L(P) \wedge \bigwedge_{\Psi \in X(P)} \lnext \Psi$ and so $w, \la\ra, z \models \Phi$, a contradiction.
  \todo{Double check \ltail}
\end{proofE}

This provides us a decidable method for verifying \iac{LTLf} property $\Phi$ against a \golog program \prog.
However, the goal is to determine a policy that executes \prog while satisfying $\Phi$.

\subsection{Synthesis}

Above, we have described a finite game arena \ts that captures the executions of a program \prog while tracking the satisfaction of a given \ac{LTLf} formula $\Phi$.
In the following, we use a game-theoretic approach on \ts to determine a policy that successfully executes \prog while satisfying $\Phi$.
We do so by defining a game between two players, the system and the environment, that play on \ts.
We start by defining a \emph{strategy}, which intuitively translates the conditions on a policy to the game arena \ts:
\begin{definition}[Strategy]
  Let \ts be the game arena for some \golog program \prog and temporal formula $\Phi$.
  Let $s \in \tsstates$ be a state of \ts.
  A set of actions $U \subseteq \actions$ is \emph{valid} in $s$ under the following conditions:
  \begin{enumerate}
    \item if $\alpha \in U$, then there is an edge $s \xrightarrow{\alpha} s'$ for some $s' \in \tsstates$
    \item if $s \xrightarrow{\alpha} s'$ for some $\alpha \in \actions_E$ and $s' \in \tsstates$, then $\alpha \in U$
    \item if $U = \emptyset$, then $s$ is a final state
  \end{enumerate}
  A \emph{strategy in \ts} is a partial function $\sigma: \tsstates \rightarrow 2^{\mathcal{A}}$ such that:
  \begin{enumerate}
    \item $\sigma$ is defined on every initial state of \ts
    \item if $\sigma$ is defined on $s \in \tsstates$, then $\sigma(s)$ is valid in $s$
    \item if $\sigma$ is defined on $s \in \tsstates$, $\alpha \in \sigma(s)$, and $s \xrightarrow{\alpha} s'$ for some $s' \in \tsstates$, then $\sigma$ is defined on $s'$
  \end{enumerate}
  We also write $s \xrightarrow{\sigma} s'$ if there is $\alpha \in \sigma(s)$ such that $s \xrightarrow{\alpha} s'$.
  A strategy $\sigma$ induces a set of plays $\plays(\sigma)$, which are those paths in \ts consistent with $\sigma$.
  Formally, $p = \la s_0, \ldots, s_n \ra \in \plays(\sigma)$ if
  \begin{enumerate}
    \item $s_0$ is an initial state of \ts
    \item for each $i$, $s_i \xrightarrow{\sigma} s_{i+1}$
    \item $\sigma(s_n) \subseteq \actions_E$ and $s_n$ is a final state of \ts
  \end{enumerate}
  A play is \emph{winning} if it ends in an accepting state. 
  A strategy $\sigma$ is \emph{winning} if every play $p \in \plays(\sigma)$ is winning.
  We call a strategy $\sigma$ \emph{terminating} if for every infinite sequence of states $s_0, s_1, \ldots$ with $s_k \xrightarrow{\sigma} s_{k+1}$ for every $k$, it holds that for every $i$, there is a $j \geq i$ such that $\sigma(s_j) \subseteq \actions_E$ and $s_j$ is final.
\end{definition}

\begin{propositionE}
  There is a terminating and winning strategy $\sigma$ in \ts if and only if there exists a terminating policy $\pi$ for \prog that
  satisfies $\Phi$.
\end{propositionE}
\begin{proofE} ~ \\
  \textbf{$\Rightarrow$:}
  Let $\sigma$ be a terminating and winning strategy in \ts.
  For a play $p = \la s_0, \ldots, s_n \ra \in \plays(\sigma)$, let $\playacts(p)$ denote the (unique) trace $\la \alpha_1, \ldots, \alpha_n \ra$ such that $s_0 \xrightarrow{\alpha_1} s_1 \xrightarrow{\alpha_2} \cdots \xrightarrow{\alpha_n} s_n$.
  We construct $\pi$ as follows:
  For every play with $p = \la s_0, \ldots, s_n \ra \in \plays(\sigma)$ where $s_i = (\tau, E_i, A_i, \rho_i)$ (note that by definition, $\tau$ is the same for each $s_i$) and $\playacts(p) = z = \la \alpha_1, \ldots, \alpha_n \ra$ and for every world $w$ with $\type(w) = \tau$, we define $\pi(w, z[..i], \rho_i) = \sigma(s_i)$.
  \\
  We first show that $\pi$ is a proper policy for \prog by showing that it satisfies the conditions of Definition~\ref{def:policy}:
  First, note that \ts contains an initial state with $s = (\tau, \emptyset, A, \delta)$ for every $w$ with $w \models \bat$ and so \ref{def:policy:init} is satisfied.
  Also, for every state $s$, $\sigma(s)$ is valid and hence \ref{def:policy:applicable} as well as \ref{def:policy:env-actions} is satisfied.
  Furthermore, by definition of the strategy, if $\alpha \in \sigma(s)$ and $s \xrightarrow{\alpha} s'$, then $\sigma$ is defined on $s'$ and so $\pi$ is defined on the corresponding $(w, z \cdot \alpha, \rho')$ and hence \ref{def:policy:defined-on-succ} is satisfied.
  Finally, again because each $\sigma(s)$ is valid, \ref{def:policy:non-blocking} is satisfied.

  Furthermore, $\pi$ is terminating and satisfies $\Phi$:
  From $\sigma$ being a terminating strategy, it directly follows that $\pi$ is terminating.
  Now, let $z \in \| \pi \|_w$ for some world $w$.
  By definition of $\pi$, there is a play $p = \la s_0, \ldots, s_n \ra \in \plays(\sigma)$ with $\playacts(p) = z$ for some $s_0 = (\tau, \emptyset, A_0, \delta)$ and with $\type(w) = \tau$.
  By Lemma~\ref{lma:transitionSystemPhiSatisfied}, there is some $(\chi, \theta) \in A_0$ such that $w, \la\ra, z \models \bigwedge_{\Psi \in \chi} \lnext \Psi$.
  By definition of \ts, there is a propositional assignment $P$ such that $X(P) = \chi$, $T(P) = \theta$, and $\{ (\psi, E) \mid \psi \in L(P) \} \subseteq \tau$.
  By Theorem~\ref{thm:effectAccumulation}, $w, z \models \bigwedge L(P)$ and so $w, \la\ra, z \models \Phi$.
  \\
  \textbf{$\Leftarrow$:}
  Let $\pi$ be a terminating policy for \prog that satisfies $\Phi$.
  Note that we cannot directly construct a strategy $\sigma$ from $\pi$ as the policy is defined on traces and hence we may have $\pi(w, z_1, \rho) \neq \pi(w, z_2, \rho)$ even if $z_1$ and $z_2$ correspond to the same state in \ts.
  Hence, we define $\sigma$ on \ts as follows:
  First, for any $w$ and $z \in \| \pi \|_w$ and every $i \leq |z|$, let $\rho_{z[..i]}$ be the remaining program after $z[..i]$, i.e., $\la \la\ra, \delta \ra \xrightarrow{w}^* \la z[..i], \rho_{z[..i]} \ra$.
  The program expression $\rho_{z[..i]}$ is well-defined because \prog is situation-determined.
  Now, suppose $s = (\tau, E, A, \rho)$ is a state of \ts, then let $Z_s^\pi$ be the set of traces from an initial state to $s$ that are compatible with $\pi$, i.e., $z = \la \alpha_1, \ldots, \alpha_n \ra \in Z_s$ if $s_0 \xrightarrow{\alpha_1} s_1 \xrightarrow{\alpha_2} \cdots \xrightarrow{\alpha_n} s$ is a path in \ts where $s_0$ is an initial state and $\alpha_{i+1} \in \pi(w, z[..i], \rho_{z[..i]})$ for some $w$ with $\type(w) = \tau$.
  If there is $z \in \| \pi \|_w$ such that $z[..i] \in Z_s^\pi$ and for all $j > i$, $z[..j] \not\in Z_s^\pi$ (i.e., $\pi$ does not return to $s$ after $z[..i]$), then we define $\sigma(s) = \pi(w, z[..i], \rho_{z[..i]})$.
  Otherwise, there must be a cycle in $\pi$ that passes through a final and accepting configuration (as otherwise $\pi$ would either be non-terminating or not satisfying $\Phi$).
  Hence, let $z \in \| \pi \|_w$ be the corresponding trace such that for some $i$, $z[..i] \in Z_s^\pi$, $\la z[..j], \rho_{z[..j]} \ra \in \final(w)$ for some $j > i$, and $w, \la\ra, z \models \Phi$ and $z[..k] \not\in Z_s^\pi$ for all $i < k < j$.
  We set $\sigma(s) = \pi(w, z[..i], \rho_{z[..i]})$ and so $\sigma$ visits a final and accepting state before visiting $s$ again.

  We first show that $\sigma$ is a proper strategy for \ts:
  Clearly, as $\pi$ is a proper policy and thus by Definition~\ref{def:policy}-\ref{def:policy:init} defined on every initial configuration, $\sigma$ is defined on every initial state of \ts.
  Second, every $\sigma(s)$ is valid, because $\pi$ satisfies \ref{def:policy:applicable}, \ref{def:policy:env-actions}, and \ref{def:policy:non-blocking} of Definition~\ref{def:policy}.
  Finally, $\sigma$ is defined on every $\sigma$-reachable state $s$, as $\sigma$ follows $\pi$ and by Definition~\ref{def:policy}-\ref{def:policy:defined-on-succ}, $\pi$ is defined on every successor configuration.

  It remains to be shown that $\sigma$ is winning and terminating.
  As \ts is finite, every infinite path must visit a state twice.
  By construction, $\sigma$ visits a final state before visiting a state $s$ again.
  Furthermore, as $\pi$ is terminating, there must be such a state with $\sigma(s) \subseteq \actions_E$ and so $\sigma$ is terminating.
  Finally, by construction, every play $p \in \plays(\sigma)$ corresponds to a trace $z \in \| \pi \|_w$ for some $w$ with $w, \la\ra, z \models \Phi$.
  Let $s = (\tau, E, A, \rho)$ be the last state of $p$.
  By Lemma~\ref{lma:transitionSystemPhiSatisfied}, there is $(\chi, \theta) \in A$ such that $w, z, \la\ra \models \bigwedge_{\Psi \in \chi} \lnext \Psi$.
  However, by the semantics of temporal formulas, this is only possible if $\chi = \emptyset$ and $\theta = \top$.
  Hence, $s$ is accepting and so every $p \in \plays(\sigma)$ is winning.
  As every play is winning, $\sigma$ is winning.
\end{proofE}

Hence, we need to determine a terminating and winning strategy in \ts.
In principle, this can be done with backward search starting in a set of \emph{good} states and then checking whether the agent can force every play to end in a good state.
However, not every final and accepting state is necessarily good, as the environment may force a play from this state that ends in a non-accepting state.  
On the other hand, every winning play must end in an accepting state, so if a strategy exists, there must be an enforceable set of final and accepting states.
Hence, we can guess which final and accepting states are enforceable and then check if there is indeed a strategy that can force every play to end in those states.

\begin{algorithm}[t]
  \small
  \caption{Computing a strategy from \ts}
  \label{alg:labeling}
  \begin{algorithmic}[1]
    \ForAll{$H \in 2^{\tsstates_F \cap \tsstates_A}$}
    \State $G \gets H$;
    $R \gets \{ s \in G \mid \Succ_E(s) = \emptyset \}$;
    $\sigma \gets \emptyset$
    \State $Q \gets \{ s \in \tsstates \mid \Succ(s) \cap G \neq \emptyset \}$
    \While{$Q \neq \emptyset$}
    \State $s \gets \Call{pop}{Q}$
    \If{$s \in \tsstates_F \setminus \tsstates_A \wedge \Succ_C(s) = \emptyset$} \label{alg:labeling:nonaccepting}
    \textbf{continue}
    \EndIf
    \If{$s \in R$}
    \textbf{continue}
    \EndIf
    \If{$\Succ_E(s) \neq \emptyset \wedge \forall s' \in \Succ_E(s): s' \in G \vee \hphantom{x}\Succ_E(s) = \emptyset \wedge \exists s' \in \Succ_C(s): s' \in G$} \label{alg:labeling:cond}
    \State $G \gets G \cup \{ s \}$;
    $R \gets R \cup \{ s \}$
    \If{$s \in \tsstates_F \cap \tsstates_A$}
    \State $\sigma(s) \gets \{ \alpha \mid \exists s' \in \Succ_E(s).\ s \xrightarrow{\alpha} s' \}$ \label{alg:labeling:finalaccepting}
    \Else
    $\: \sigma(s) \gets \{ \alpha \mid \exists s' \in G.\  s \xrightarrow{\alpha} s' \}$
    \EndIf
    \State $Q \gets Q \cup \{ s' \mid s \in \Succ(s') \}$
    \EndIf
    \EndWhile
    \If{$H \cup \tsstates_0 \subseteq R$}
    \Return $\sigma$
    \EndIf
    \EndFor
  \end{algorithmic}
\end{algorithm}

This approach is formalized in Algorithm~\ref{alg:labeling}.
It starts with a hypothesis $H \subseteq \tsstates_F \cap \tsstates_A$ of good states $G$ and tracks the states $R$ that can reach $G$.
It then iteratively checks the predecessors of all states in $G$ whether the agent can force the play to end in $G$.
This is the case if all environment successors $\Succ_E(s)$ are in $G$ or if there is a control successor $\Succ_C(s)$ in $G$.
If a state is found that can be forced to end in $G$, it is added to $G$ and $R$ and $\sigma$ are updated accordingly.
Finally, if all states of $H$ and all initial states $\tsstates_0$ can in fact reach $G$, then $\sigma$ is a winning and terminating strategy:
\begin{theoremE}
  Algorithm~\ref{alg:labeling} terminates and returns a winning and terminating strategy if one exists.
\end{theoremE}
\begin{proofE}
  It is easy to see that Algorithm~\ref{alg:labeling} terminates:
  Note that a state $s$ is only added to $Q$ if one of its successors is added to $R$ or if it is in $Q$ initially.
  As there are only finitely many states in \tsstates, only finitely many states can be added to $Q$, and hence $Q$ is eventually empty.
  Finally, again because \tsstates is finite, there can only be finitely many hypotheses $H$.
  \\
  We continue by showing each returned strategy is winning and terminating:
  Assume Algorithm~\ref{alg:labeling} returns a strategy $\sigma$ that is not winning.
  Then there is a play $p = \la s_0, s_1, \ldots, s_n \ra \in \plays(\sigma)$ that is not winning, i.e., ending in a state $s_n$ that is final but not accepting.
  Clearly, $s_n$ is only added to $R$ if every environment successor is in $G$, or if there is a control successor in $G$.
  As the play ends in $s_n$, $\sigma(s_n) \subseteq \actions_E$ and so every environment successor of $s_n$ is in $G$.
  However, as $s_n$ is final but non-accepting, by line~\ref{alg:labeling:nonaccepting}, $s_n$ is not added to $G$ and hence also not to $R$, contradicting the assumption.
  \\
  Now, assume $\sigma$ is non-terminating.
  Then there is an infinite sequence of $\sigma$-compatible states $s_0, s_1, \ldots$ such that for some $i$, every state $s_j$ for $j \geq i$ is non-final or $\sigma(s_j) \cap \actions_C \neq \emptyset$.
  As initially $G$ only consists of final and accepting states, it is easy to see that for every $j$, $\alpha \in \sigma(s_j)$ and $s_j \xrightarrow{\alpha} s_{j+1}$ implies that $s_{j+1}$ is closer to some final and accepting state than $s_j$.
  As there are only finitely many states in \tsstates, for every $j$, there must be a $k \geq j$ such that $s_k$ is final and accepting.
  Finally, by line~\ref{alg:labeling:finalaccepting}, $\sigma(s_k) \subseteq \actions_E$, contradicting the assumption.
  \\
  Finally, we show that the algorithm is complete.
  Assume $\sigma$ is a winning and terminating strategy but Algorithm~\ref{alg:labeling} does not return a winning and terminating strategy.
  First, from above, it directly follows that it returns $\bot$ (as any strategy returned is in fact winning and terminating).
  Now, let $H$ be the final and accepting states that are visited by $\sigma$.
  We define a distance $d(s)$ as the maximal number of steps to reach a final and accepting state from $s$ in any play of $\sigma$, i.e., $d(s) = \max \{ j \mid p_0, \ldots, p_i, s, s_1, \ldots s_j \in \plays(\sigma), s_j \in H, \forall i < j: s_i \not\in H \}$.
  Clearly, $d(s)$ is defined and finite for all initial states $s$ and all states in $H$, as otherwise $\sigma$ would not be winning.
  We can now show by induction on $d(s)$ that every state $s$ visited by $\sigma$ is added to $G$.
  The base case is trivial.
  For the induction step, let $s$ be a state with $d(s) = n$ and assume that every state $s'$ with $d(s') < n$ is in $G$.
  As $\sigma$ is winning, for every $s \in \Succ_E(s)$, there is an action $\alpha \in \sigma(s)$ such that $s \xrightarrow{\alpha} s'$.
  By definition, $d(s') < n$ and so $s' \in G$.
  If $\Succ_E(s) = \emptyset$, there must be an action $\alpha \in \actions_C$ with $\alpha \in \sigma(s)$.
  Again, for every $s' \in \Succ_C(s)$, $d(s') < n$ and so $s' \in G$.
  By line~\ref{alg:labeling:cond}, $s$ is added to $G$.
  Hence, after the while loop terminates, $H \cup \tsstates \subseteq R$ and so the algorithm returns some strategy, a contradiction.
\end{proofE}

\section{Experiments}
\label{sec:evaluation}

We implemented the method \cite{classen_2024_14526973} in the Prolog-based Golog interpreter \vergo~\cite{DBLP:conf/kr/Classen18},
that, different from other implementations, uses full FOL as base logic,
where an embedded theorem prover \cite{DBLP:conf/cade/0001CV19} is used for reasoning tasks such as deciding entailment and consistency.
The system contains optimizations for handling FO expressions, in particular an FO variant of binary decision diagrams.
We evaluated the method on two domains, a dishwasher robot and a warehouse robot, that we will be described in detail below.
All experiments were conducted on an Intel\textregistered\ Core\texttrademark\ i5-7300U @2.60GHz with 8GB of RAM,
running Debian 12 with WSL2 under Windows 10, using SWI-Prolog 9.0.4 and version 3.2 of the E theorem prover.

\subsection{Incremental Construction}

In our implementation, the construction of the abstract game arena follows closely Definition \ref{def:TransitionSystem}.
However, the construction is done in an incremental fashion, where only the relevant and reachable parts are actually materialized.
This is achieved by keeping the types as general as possible, and only including additional formulas once they are needed.
More specifically, the method works by iterating the following steps, until no more changes occur:
\begin{description}[leftmargin=0pt]
\item[Initialize:] Create initial states $(\tau, \emptyset, A, \delta)$, where types $\tau$ are constructed only from formulas in $\bat_0$ and literals $L(P)$ of propositional assignments over $A$.
\item[Split:] If there is a state $(\tau, E, A, \rho)$ that does not entail a truth value for some required condition $\psi$ (the transition condition for an action $\alpha$, the termination condition $\termcond{\rho}$, the condition $\contcon$ of an effect, or a literal $l \in L(P)$ of a propositional assignment over $A$), then create two copies of all states and transitions, where one includes $\psi$ and the other includes $\lnot\psi$ into $\tau$, discarding states with inconsistent $\tau$.
\item[Expand:] If a state $s = (\tau, E, A, \rho)$ admits an action $\alpha$, create the successor state $s'$ and the transition $s \xrightarrow{\alpha} s'$.
\end{description}
We represent $\tau$ directly by the regressed versions of formulas to avoid having to regress them repeatedly.
The construction also stops in states where $A = \emptyset$, since the corresponding traces can never satisfy the input property.






\subsection{Dishwasher Robot}

The first domain is inspired by the dishwasher robot example used in \cite{classenExploringBoundariesDecidable2014}, but adds additional fluents.
A robot can move between a number of rooms and the kitchen, load (an arbitrary number of) dirty dishes onto itself, and unload dishes it carries into the dishwasher.
The environment has actions that represent used dishes being placed in arbitrary rooms.
Every dish can only be used once in this fashion.
The basic action theory, program, and temporal specification are specified below.

\newcommand*{\rat}{\rfluent{at}}

\paragraph{Initial situation:}
\begin{sizeddisplay}
\begin{align*}
  &\rfluent{dish}(x) \equiv (x = d_1 \vee x = d_2),~ \rfluent{room}(x) \equiv (x = r_1 \vee x = r_2)
  \\
  &\forall x.\, \rat(x) \equiv x = \mi{kitchen}
  \\
  &\forall x.\, \rfluent{new}(x) \equiv \rfluent{dish}(x)
  \wedge \forall y.\, \neg \rfluent{dirtyDish}(x, y) \wedge \neg \rfluent{onRobot}(x)
  \\
  &\rfluent{onRobot}(x) \supset \rfluent{dish}(x) \wedge \neg \exists y \rfluent{dirtyDish}(x, y)
  \\
  &\rfluent{dirtyDish}(x, y) \supset \rfluent{dish}(x) \wedge \rfluent{room}(y) \wedge \neg \rfluent{onRobot}(x)
\end{align*}
\end{sizeddisplay}

\paragraph{Precondition axioms:}
\begin{sizeddisplay}
\begin{align*}
  \square&\poss(\action{load}{x, y}) \equiv \rfluent{dirtyDish}(x,y) \wedge \rat(y)\\
  \square&\poss(\action{unload}{x}) \equiv \rfluent{onRobot}(x) \wedge \rat(\mi{kitchen})\\
  \square&\poss(\action{addDish}{x, y}) \equiv \rfluent{new}(x) \wedge \rfluent{room}(y)\\
  \square&\poss(\action{goto}{x}) \equiv \rfluent{room}(x) \vee x = \mi{kitchen}
\end{align*}
\end{sizeddisplay}

\paragraph{Successor state axioms:}
\begin{sizeddisplay}
\begin{align*}
  \square [a] & \rfluent{dirtyDish}(x, y) \equiv a = \action{addDish}{x, y} \\ &\quad \vee \rfluent{dirtyDish}(x, y) \wedge a \neq \action{load}{x, y}
  \\
  \square [a] & \rfluent{onRobot}(x) \equiv \exists y.\, a = \action{load}{x, y}
  \\ &
  \quad \vee  \rfluent{onRobot}(x) \wedge a \neq \action{unload}{x}
  \\
  \square [a] & \rfluent{new}(x) \equiv \rfluent{new}(x) \wedge \neg \exists y.\, a = \action{addDish}{x, y}
  \\
  \square [a] & \rat(x) \equiv a = \action{goto}{x}
  \vee \rat(x) \wedge \neg \exists y. a = \action{goto}{y}
\end{align*}
\end{sizeddisplay}

\paragraph{Program:}

The program is shown in Algorithm~\ref{alg:dishwasher}.
It is to be understood as being {\em precondition extended},
i.e., an underlined action $\underline{A(\vec{o})}$ stands for $\pi?;A(\vec{o})$,
where $\pi$ is the right-hand side of the precondition axiom for $A$, instantiated by $\vec{o}$.
For better readability, $\iter{\delta}$ is written as $\textbf{loop}~\delta$.

\begin{algorithm}[tb]
  \small
  \caption{The program for the dishwasher robot}
  \label{alg:dishwasher}
  \begin{algorithmic}
    \Loop
    \While{$\exists x.\, \rfluent{onRobot}(x)$}
    $\pi x : \{ d_1, d_2 \}.\, \paction{unload}{x}$
    \EndWhile
    \State $\pi y: \{ r_1, r_2 \}.\, \paction{goto}{y}$;
    \While{$\exists x.\, \rfluent{dirtyDish}(x, y)$}
    $\pi x: \{ d_1, d_2\}.\, \paction{load}{x, y}$
    \EndWhile
    \State $\paction{goto}{\mi{kitchen}}$
    \EndLoop
    \State $\|$
    \Loop
    $\: \pi x: \{ d_1, d_2 \}, y: \{ r_1, r_2\}.\, \paction{addDish}{x, y}$
    \EndLoop
  \end{algorithmic}
\end{algorithm}

\paragraph{Specification:}
$
  \lfin \lglob \neg \exists x,y.\, \rfluent{dirtyDish}(x,y)
$

\begin{table}
\centering
\small
\setlength{\tabcolsep}{2pt}
\begin{filecontents*}{results_dish.csv}
12/18/2024 05:10:01 AM,1,1,main,prop1,22,25,16,19,2567,1794
12/18/2024 05:10:04 AM,1,2,main,prop1,150,203,128,176,155412,149566
12/18/2024 05:12:39 AM,1,3,main,prop1,n/a,n/a,n/a,n/a,timeout,timeout
12/18/2024 06:12:40 AM,2,1,main,prop1,109,168,87,110,69021,66700
12/18/2024 06:13:49 AM,2,2,main,prop1,n/a,n/a,n/a,n/a,timeout,timeout
12/18/2024 08:13:50 AM,3,1,main,prop1,483,857,413,543,1885719,1879396
12/18/2024 08:45:16 AM,3,2,main,prop1,n/a,n/a,n/a,n/a,timeout,timeout
\end{filecontents*}
\DTLloaddb[noheader,keys={time,rooms,dishes,program,property,nodes,edges,stnodes,stedges,twc,tcc}]{dishrobot}{results_dish.csv}
\begin{tabular}{ccccccc}
  R & D & Nodes (TS) & Edges (TS) & Nodes (St) & Edges (St) & Time [s]
  \tabularnewline\hline%
    \DTLforeach{dishrobot}{%
      \rooms=rooms, \dishes=dishes, \nodes=nodes, \edges=edges,
      \twc=twc, \stnodes=stnodes, \stedges=stedges}{%
      \DTLiffirstrow{}{\tabularnewline}%
            \rooms & \dishes &
            \DTLsubstitute{\nodes}{n/a}{--}\nodes &
            \DTLsubstitute{\edges}{n/a}{--}\edges &
            \DTLsubstitute{\stnodes}{n/a}{--}\stnodes &
            \DTLsubstitute{\stedges}{n/a}{--}\stedges &
            \DTLifstringeq{\twc}{timeout}{--}{\FPeval{\twcRounded}{round(\twc/1000,1)}\twcRounded}
            \DTLiflastrow{}{}
    }
\end{tabular}
\caption{Evaluation Results for the Dish Robot Domain}
\label{tab:dishes}
\end{table}

\paragraph{Results:}
Table \ref{tab:dishes} summarizes the experimental results on the dishwasher domain.
Here, \emph{R} and \emph{D} denote the number of rooms and dishes, respectively,
\emph{Nodes (TS)} and \emph{Edges (TS)} refer to the number of nodes and edges of the resulting transition system,
while \emph{Nodes (St)} and \emph{Edges (St)} denote the corresponding metrics of the discovered strategy.
\emph{Time} indicates the duration in seconds for completing the algorithm, with a timeout set at 120 minutes.
As expected, the size of transition system, and the time needed to construct it, grows with additional rooms or dishes.
Interestingly, the number of dishes has a bigger impact than the number of rooms.
Intuitively, this is because the program contains more choices for dishes than for rooms, which are furthermore nested inside inner loops.
Accordingly, adding one more dish results in a more significant blow-up than adding a room.






\subsection{Warehouse Robot}

The second domain is a warehouse robot, adapted from an example in \cite{classenDecidableVerificationDecisionTheoretic2017}.
Here, the robot can move boxes from one shelf of a warehouse to another.
The boxes may contain an unknown number of objects, and it is unknown whether and which objects are fragile.
Accidentally (i.e., due to the environment's choice), the robot may drop a box, breaking all fragile objects in it, unless the box contains bubble wrap.
The robot has the option to put bubble wrap into a box.

\renewcommand*{\rat}{\rfluent{rAt}}
\newcommand*{\at}{\rfluent{at}}

\paragraph{Initial situation:}
\begin{sizeddisplay}
\begin{align*}
  &\forall x.\, \rfluent{shelf}(x) \equiv (x = s_1 \vee x = s_2)
  \\
  &\forall x.\, \rfluent{box}(x) \equiv (x = b_1 \vee x = b_2)
  \\
  &\forall x.\, \exists y \rfluent{in}(x, y) \supset \neg \rfluent{shelf}(x) \wedge \neg \rfluent{box}(x)
  \\
  &\exists x.\, \rfluent{wrap}(x)
  \\
  &\forall x.\, \neg \rfluent{broken}(x) \wedge \neg \rfluent{holding}(x)
  \\
  &\rat(s_1) \wedge \forall x.\, \rfluent{box}(x) \supset \at(x, s_1)
  \\
  &\forall x, y.\, (\rfluent{in}(x,y) \wedge \rfluent{box}(y)) \supset \at(x, s_1)
  \\
  &\forall y.\, y \neq s_1 \supset \neg \rat(y) \wedge \forall x.\, \neg \at(x, y)
  \\
  &
  \forall x, y.\, \rfluent{in}(x,y) \supset \neg \rfluent{wrap}(x)
\end{align*}
\end{sizeddisplay}

\paragraph{Precondition axioms:}
\begin{sizeddisplay}
\begin{align*}
  \square&\poss(\action{take}{x, y}) \equiv \at(x, y) \wedge \rat(y)\\
  \square&\poss(\action{move}{x, y}) \equiv \rat(x) \wedge \rfluent{shelf}(y) \wedge (x \neq y)\\
  \square&\poss(\action{put}{x, y}) \equiv \rfluent{holding}(x) \wedge \rat(y)\\
  \square&\poss(\action{addWrap}{x}) \equiv \exists y.\, \rat(y) \wedge \at(x, y)\\
  \square&\poss(\action{drop}{x}) \equiv \rfluent{holding}(x)
\end{align*}
\end{sizeddisplay}

\paragraph{Successor state axioms:}
\begin{sizeddisplay}
\begin{align*}
  \square [a] & \rat(y) \equiv \exists x.\, a = \action{move}{x, y} \\ &\qquad \vee \rat(y) \wedge \neg \exists z. a = \action{move}{y, z}
  \\
  \square [a] & \at(x, y) \equiv \exists z [a = \action{move}{z, y}~\wedge\\ & \exists v (\rfluent{holding}(v) \wedge (v = x \vee \rfluent{in}(x,v)))] \\ & \vee \at(x, y) \wedge \neg \exists z [a = \action{move}{y, z}~\wedge\\ & \exists v (\rfluent{holding}(v) \wedge (v = x \vee \rfluent{in}(x,v)))]
  \\
  \square [a] & \rfluent{holding}(x) \equiv \exists y.\, a = \action{take}{x, y} \\ &\qquad \vee \rfluent{holding}(x) \wedge \neg \exists y.\, a = \action{put}{x, y}
  \\
  \square [a] & \rfluent{broken}(x) \equiv \exists y.\, a = \action{drop}{y} \wedge \rfluent{in}(x, y) \wedge \rfluent{fragile}(x)
  \\
              & \wedge \neg \exists z.\ \rfluent{in}(z, y) \wedge \rfluent{wrap}(z) \vee \rfluent{broken}(x)
  \\
  \square [a] & \rfluent{in}(x, y) \equiv a = \action{addWrap}{y} \wedge \rfluent{wrap}(x) \vee \rfluent{in}(x, y)
\end{align*}
\end{sizeddisplay}

\begin{algorithm}[htb]
  \small
  \caption{The program for the warehouse robot}
  \label{alg:warehouse}
  \begin{algorithmic}
    \Loop
    \State $\pi l_0, l_1 : \{ s_1, s_2 \}.\, \big[ \paction{move}{l_0, l_1}^?;$
    \State $\quad \pi b \;\mathbf{:}\; \{ b_1, b_2 \}.\, \big( \paction{wrap}{b}^?; \paction{take}{b, s_1}; \paction{drop}{b}^?;$
    \State $\quad\quad \pi l_2 : \{ s_1, s_2 \}.\, \paction{move}{l_1, l_2}; \paction{put}{b, l_2}\big)\big]$
    \EndLoop
  \end{algorithmic}
\end{algorithm}



\paragraph{Program:}
The program for the warehouse robot is shown in Algorithm~\ref{alg:warehouse}.
The notation $\delta^?$ stands for an optional execution of $\delta$, and is formally defined as $\delta^? \eqdef (\delta \mid \nil)$.
Note that the choice for putting bubble wrap is up to the robot, but that of the box getting dropped is due to the environment.

\paragraph{Specification:}
  $\lfin \forall o.\, \rfluent{in}(o, b_1) \supset \neg \rfluent{broken}(o) \wedge \at(o, s_2)$




\begin{table}
\centering
\small
\setlength{\tabcolsep}{2pt}
\begin{filecontents*}{results_warehouse.csv}
12/18/2024 02:15:57 PM,2,1,main,prop2,162,162,46,58,11834,6819
12/18/2024 02:16:09 PM,2,2,main,prop2,7584,7830,2038,2647,13328897,12932938
12/18/2024 05:58:17 PM,2,3,main,prop2,n/a,n/a,n/a,n/a,timeout,timeout
\end{filecontents*}
\DTLloaddb[noheader,keys={time,shelves,boxes,program,property,nodes,edges,stnodes,stedges,twc,tcc}]{warehouserobot}{results_warehouse.csv}
\begin{tabular}{ccccccc}
  B & Nodes (TS) & Edges (TS) & Nodes (St) & Edges (St) & Time [s]
  \tabularnewline\hline%
    \DTLforeach{warehouserobot}{%
      \boxes=boxes, \nodes=nodes, \edges=edges,
      \twc=twc, \stnodes=stnodes, \stedges=stedges}{%
      \DTLiffirstrow{}{\tabularnewline}%
            \boxes &
            \DTLsubstitute{\nodes}{n/a}{--}\nodes &
            \DTLsubstitute{\edges}{n/a}{--}\edges &
            \DTLsubstitute{\stnodes}{n/a}{--}\stnodes &
            \DTLsubstitute{\stedges}{n/a}{--}\stedges &
            \DTLifstringeq{\twc}{timeout}{--}{\FPeval{\twcRounded}{round(\twc/1000,1)}\twcRounded}
            \DTLiflastrow{}{}
    }
\end{tabular}
\caption{Evaluation Results for the Warehouse Domain}
\label{tab:warehouse}
\end{table}

\paragraph{Results:}
Table \ref{tab:warehouse} presents the results of the experiments on the warehouse robot domain,
where B denotes the number of boxes,
and the other columns are as before (with a timeout of 240 minutes).
As can be seen, the method struggles more with this domain than the previous one, which is due to several reasons.
For one, the successor state axioms for the warehouse robot actually exploit the expressivity of the class of acyclic theories more than do the ones for the dishwasher robot.
Note that the dishwasher BAT actually falls into the class of {\em local-effect} theories \cite{DBLP:conf/ijcai/LiuL09}, a subset of acyclic theories where regression works much simpler (i.e., does not introduce additional quantifiers), and consequently results in less complex formulas.
Moreover, the warehouse robot suffers from the same problem that causes the {\em Gripper} domain to be a challenge in classical planning:
There is a number of objects, each of which has to be handled in the same way.
For solving the task, the order in which objects are handled is hence irrelevant, yet the system considers all possible permutations, resulting in a blow-up.
The problem is amplified by the fact that handling a single box in this domain is a slightly complex task in itself, containing a sequence of actions with several choice points.
An interesting avenue for future work would be to improve our method to be able to detect and deal with symmetries of this kind.

\section{Conclusion}
\label{sec:conclusion}
In this paper, we have presented an approach to the realization of \golog programs with uncontrollable actions.
We have formulated the realization problem as a synthesis problem, where parts of the program  are under the environment's control and the agent needs to determine a policy that realizes the program while satisfying the temporal specification.
The presented approach synthesizes policies for \ac{LTLf} specifications on \golog programs with first-order action theories that allow for an unbounded number of objects and non-local effects, an expressive and decidable fragment of the situation calculus.
We have demonstrated the feasibility of the approach in two example domains. 
The synthesis method can also be understood as a (restricted) first-order variant of \ac{LTLf} synthesis, where the user may provide a declarative specification of the agent's capabilities along with a partial strategy.
Future work could further investigate this relation.

\section*{Acknowledgements}
The research has been supported by the Alexander von Humboldt Foundation with funds from the Federal Ministry for Education and Research, Germany,  by the European Research Council (ERC), Grant agreement No.~885107, and by the Excellence Strategy of the Federal Government and the L\"{a}nder, Germany.

\bibliography{zotero,references}

\ifthenelse{\boolean{reprochecklist}}{
\section*{Reproducibility Checklist}

\begin{itemize}
\item This paper:
\begin{itemize}
\item Includes a conceptual outline and/or pseudocode description of AI methods introduced: yes
\item Clearly delineates statements that are opinions, hypothesis, and speculation from objective facts and results:yes
\item Provides well marked pedagogical references for less-familiare readers to gain background necessary to replicate the paper: yes
\end{itemize}

\item Does this paper make theoretical contributions? yes
\begin{itemize}
\item All assumptions and restrictions are stated clearly and formally: yes
\item All novel claims are stated formally (e.g., in theorem statements): yes
\item Proofs of all novel claims are included: yes
\item Proof sketches or intuitions are given for complex and/or novel results: yes
\item Appropriate citations to theoretical tools used are given: yes
\item All theoretical claims are demonstrated empirically to hold: NA
\item All experimental code used to eliminate or disprove claims is included: NA
\end{itemize}

\item Does this paper rely on one or more datasets? no
\item Does this paper include computational experiments? yes
\begin{itemize}
\item Any code required for pre-processing data is included in the appendix. yes
\item All source code required for conducting and analyzing the experiments is included in a code appendix. yes
\item All source code required for conducting and analyzing the experiments will be made publicly available upon publication of the paper with a license that allows free usage for research purposes. yes
\item All source code implementing new methods have comments detailing the implementation, with references to the paper where each step comes from yes
\item If an algorithm depends on randomness, then the method used for setting seeds is described in a way sufficient to allow replication of results. NA
\item This paper specifies the computing infrastructure used for running experiments (hardware and software), including GPU/CPU models; amount of memory; operating system; names and versions of relevant software libraries and frameworks. yes
\item This paper formally describes evaluation metrics used and explains the motivation for choosing these metrics. yes
\item This paper states the number of algorithm runs used to compute each reported result. yes
\item Analysis of experiments goes beyond single-dimensional summaries of performance (e.g., average; median) to include measures of variation, confidence, or other distributional information. NA
\item The significance of any improvement or decrease in performance is judged using appropriate statistical tests (e.g., Wilcoxon signed-rank). NA
\item This paper lists all final (hyper-)parameters used for each model/algorithm in the paper’s experiments. NA
\item This paper states the number and range of values tried per (hyper-) parameter during development of the paper, along with the criterion used for selecting the final parameter setting. NA
\end{itemize}
\end{itemize}
}{}

\ifthenelse{\boolean{techreport}}{
\clearpage
\appendix
\section{Proofs}
\printProofs
}{}

\end{document}